\pdfoutput=1

\documentclass[11pt]{article}

\usepackage{acl}

\usepackage{times}
\usepackage{latexsym}
\usepackage{amsfonts}
\usepackage{subfigure}
\usepackage{comment}

\usepackage[T1]{fontenc}

\usepackage[utf8]{inputenc}

\usepackage{microtype}

\usepackage{inconsolata}

\usepackage{graphicx}

\title{Sci-LoRA: Mixture of Scientific LoRAs for Cross-Domain Lay Paraphrasing}

\author{\bf Ming Cheng, Jiaying Gong, Hoda Eldardiry \\
Virginia Tech \\
\texttt{\{ming98,gjiaying,hdardiry\}@vt.edu}
}

\begin{document}
\maketitle
\begin{abstract}
Lay paraphrasing aims to make scientific information accessible to audiences without technical backgrounds.
However, most existing studies focus on a single domain, such as biomedicine. 
With the rise of interdisciplinary research, it is increasingly necessary to comprehend knowledge spanning multiple technical fields.
To address this, we propose Sci-LoRA, a model that leverages a mixture of LoRAs fine-tuned on multiple scientific domains.
In particular, Sci-LoRA dynamically generates and applies weights for each LoRA, enabling it to adjust the impact of different domains based on the input text, without requiring explicit domain labels.
To balance domain-specific knowledge and generalization across various domains, Sci-LoRA integrates information at both the data and model levels. This dynamic fusion enhances the adaptability and performance across various domains.
Experimental results across twelve domains on five public datasets show that Sci-LoRA significantly outperforms state-of-the-art large language models and demonstrates flexible generalization and adaptability in cross-domain lay paraphrasing.
\end{abstract}

\section{Introduction}
Lay paraphrasing aims at making the technical or specialist text comprehensible for non-expert audiences.
In an era of abundant specialized knolwedge, lay paraphrasing plays a crucial role in making intricate scientific information and concepts accessible and understandable to people lack of technical expertise, fostering public understanding of science and its impact.
With the growing prevalence of interdisciplinary research and collaboration, scientific contents from diverse domains is increasingly reaching the general public.
It is particularly important to make the interdisciplinary information comprehensible to the cross-domain non-expert audience for minimizing misunderstandings and fostering effective cross-domain collaboration.
Table~\ref{tab:example} presents examples of interdisciplinary research content along with their layman-friendly paraphrased versions. For example, the term "influenza A/H1N1 hemagglutinin (HA)" can be simplified to "flu virus" for audiences without a biology background. Similarly, the mathematical expression of a graph, "$G = (V, E)$", where $V$ represents vertices, can be rephrased as "each point represents a version of the virus" for those unfamiliar with computer science concepts.


\begin{table*}[]
\small
\centering
\caption{Examples of interdisciplinary scientific contents and lay paraphrasing for the general audience.}
\label{tab:example}
\begin{tabular}{|p{7.5cm}|p{7.5cm}|} 
\hline
\textbf{Technical Contents} & \textbf{Lay Paraphrasing} \\ \hline
\textbf{CS + Art History}: We propose a four-step human-AI collaboration workflow to support the discovery and clustering of these backdrops. Focusing on the painted backdrops of the American Civil War, we present Backdrop Explorer, a content-based image retrieval (CBIR) system incorporating computer vision and novel user interactions. & We developed a system called Backdrop Explorer to help people find and organize images of painted backdrops from the American Civil War. It uses advanced computer technology to analyze and group similar images. The process involves four steps where humans and artificial intelligence work together to make searching easier and more interactive. \\ \hline
\textbf{CS + Computational Biology}: We suggest representing antigenic drifts within influenza A/H1N1 hemagglutinin (HA) protein as a graph, $G = (V, E)$, where $V$ is the set of vertices representing each possible sequence and $E$ is the set of edges representing single amino acid substitutions.   &  We suggest using a network-like diagram to track how the flu virus changes over time. In this diagram, each point represents a version of the virus, and lines between them show small changes in the virus's building blocks.  \\ \hline
\textbf{CS + Chemistry}: Constructing a robust deep learning model for assessing materials' structure-property relationships remains a non-trivial task due to highly flexible model architecture and the challenge of selecting appropriate material representation methods. In this regard, we develop advanced deep-learning models and implement them for predicting the quantum-chemical calculated properties (i.e., formation energy) for an enormous number of crystal systems. & Building a strong deep learning model to understand the relationship between the structure of materials and their properties is still a difficult task. This is because the models can be very complex, and it’s not always easy to choose the best way to represent the materials. To tackle this, we’ve developed advanced deep learning models that can predict certain properties of materials, like formation energy, for a large number of crystal systems. \\ \hline
\end{tabular}
\end{table*}

However, existing works on lay paraphrasing are limited to a single domain, such as biomedicine~\cite{guo2024retrieval, fonseca-cohen-2024-large-language}, scientific news~\cite{liu-etal-2024-p3sum}, etc. 
Though the most recent study has expanded the scope of lay paraphrasing to translate technical language into general-audience language across multiple domains~\cite{cheng-etal-2025-vtechagp}, the models are still only fine-tuned on separate domain-specific data, assuming domain knowledge is available during inference and not fully leveraging the cross-domain knowledge. 
These studies focus on developing one model within a specified domain while overlooking the generalization of the model across multiple domains, resulting in potential misinterpretations of cross-domain concepts. 
Besides, it is hard for those models to adapt to unseen information when new scientific interdisciplinary fields emerge. They often require full-scale retraining, which is inefficient and time-consuming.


To address the above challenges, this paper explores the "mixed-domain scenario", where scientific content may span one or multiple domains.
We introduce Sci-LoRA, a model that leverages a mixture of LoRAs fine-tuned on a diverse set of scientific domains.
Unlike conventional models that are fine-tuned for one specific domain, Sci-LoRA adopts a multi-LoRA serving architecture, enabling continuous improvement as new LoRA modules can be added and updated for newly emerging domains.
To effectively utilize cross-domain knowledge, an adapter weight generator is designed to dynamically generate and assign weights to each LoRA, adjusting the influence based on the relevance of different domains.
Specifically, a text encoder is trained using contrastive learning to better distinguish representations across domains.
Then, weights are calculated based on the similarity between input text and domain adapter representations.
To further enhance cross-domain generalization, a dynamic LoRA fusion module is employed to integrate domain-specific knowledge from mixture of LoRAs with generalized information from the mixture of data.
This approach allows Sci-LoRA to effectively generalize across domains while maintaining domain-specific accuracy.
Sci-LoRA significantly broadens access to cross-domain scientific content for non-expert audiences and facilitates interdisciplinary research collaboration.
Contributions are summarized as:
\begin{itemize}
    \item We propose Sci-LoRA, a model that leverages a mixture of LoRAs fine-tuned on multiple scientific domains, designed for the automatic cross-domain lay paraphrasing task.
    \item We design the adapter weight generator and dynamic LoRA fusion that generate adaptive weights and integrate domain-specific knowledge with generalizd information.
    \item Extensive experiments over ten evaluation metrics on five public datasets across twelve different domains demonstrate the superior effectiveness and generalization capability of Sci-LoRA over state-of-the-art models.  
\end{itemize}

\section{Related Work}
\subsection{Lay Paraphrasing}
Lay paraphrasing focuses on rewriting the text written from the technical experts to the general audience without specialized domain knowledge~\cite{cheng-etal-2025-vtechagp, guo2024retrieval}. 
Recent methods on lay summarization or lay paraphrasing tasks are restricted to a single domain.
For example, large language models are finetuned and deployed to generate lay/plain language text in the biomedicine domain~\cite{guo2024retrieval, fonseca-cohen-2024-large-language, attal2023dataset, tang-etal-2023-improving, guo2021automated}, science and engineering domain~\cite{cardenas-etal-2023-dont}, news domain~\cite{liu-etal-2024-p3sum}, and literature domain~\cite{pu-etal-2023-incorporating}.
These studies focus on developing one model within a specified domain while overlooking the generalization of the model for multiple domains.
Besides, these works mainly focus on summarization instead of paraphraph paraphrasing.
Given the imbalanced data distribution across various domains~\cite{cheng-etal-2025-vtechagp} and the growing prominence of interdisciplinary research fields, we aim to develop an effective and efficient method that performs well and demonstrates robustness for the cross-domain lay paraphrasing task.

\subsection{Mixture of Loras}
As the parameter scale increases for large language models (LLMs), parameter-efficient fine-tuning (PEFT)~\cite{houlsby2019parameter, han2024parameter} strategies like Low-Rank Adaptation (LoRA)~\cite{hu2022lora} efficiently adapting LLMs to multiple data domains by fine-tuning a small subset of parameters.
Motivated by the multisource domain adaptation of Mixture of Experts (MoE)~\cite{cai2024survey, guo-etal-2018-multi}, which is an ensemble method with a combination of sub-modules or experts for improving the overall performance in diverse tasks, mixture of LoRAs combines multiple low-rank modules to enhance adaptability and performance for more efficient and effective model tuning in different data domains.

Recent works on mixture of LoRAs can be roughly divided into three categories: (1) linear merge, where different LoRAs have the same static weights~\cite{yadav2024ties, yu2024language, huang2023lorahub}; (2) router, additional linear layers for LoRA selection or rule-based LoRA selection~\cite{feng-etal-2024-mixture, zhang2024dlp, zhao2024mosld, liu2024learning, muqeeth2024learning, 10.1063/5.0203126}; and (3) trainable gating networks, modeling the optimal distribution of combination weights for various LoRAs~\cite{wu2024mixture, xu2024meteora, prabhakar2024lora, zhao2024merging, luo2024moelora, zhao-etal-2024-loraretriever, lv-etal-2024-hyperlora}.
For the lay paraphrasing or summarization task, studies leverage LoRA for efficient fine-tuning on biomedical articles~\cite{malik-etal-2024-hgp, kim2024saama}.
However, these works only focus on LoRA tuning in a specific biomedical domain.
Motivated by mixture of LoRAs, we focus on generalization for cross-domain lay paraphrasing based on mixture of scientific LoRAs.


\section{Methodology}
\subsection{Background}
\subsubsection{Low-Rank Adaptation}
Low-Rank Adaptation (LoRA)~\cite{hu2022lora} is a parameter-efficient fine-tuning method designed for finetuning LLMs by integrating trainable low-rank matrices instead of updating all parameters of LLMs during training.
Consider a weight matrix $W \in \mathbb{R}^{d_{in} \times d_{out}}$ within the original LLMs, where $d_{in}$ and $d_{out}$ are input and output dimensions, respectively.
LoRA injects two low-rank matrices, $A \in \mathbb{R}^{d_{in} \times r}$ and $B \in \mathbb{R}^{r \times d_{out}}$, where the rank $r \ll min(d_{in}, d_{out})$.
Instead of directly updating $W$, LoRA modifies model's 
forward process for one layer as the following:
\begin{equation}
f(x) = Wx + \triangle Wx = Wx + BAx
\end{equation}
where $x \in \mathbb{R}^{d_{in}}$ denotes the input.

\subsubsection{Problem Formulation}
Lay paraphrasing is the process of rephrasing scientific or technical language into simpler, and more accessible language that can be easily understood by a general audience. 
Let $D_{multi} =  \left\{D_{1}, D_{2}, \dots, D_{n} \right\}$ represent $n$ different techical domains data, where $D_{i} = \left\{X_{i}, Y_{i} \right\}$.
$X_{i}$ is the technical text (inputs) and $Y_{i}$ is the general-audience text (outputs). $X_{i} = \left\{x_{1}, x_{2}, \cdots, x_{m} \right\}$, $Y_{i} = \left\{y_{1}, y_{2}, \cdots, y_{m} \right\}$, where $m$ is the number of textual documents.
Given a LLM $\mathcal{M}$, and a set of LoRA adapters $\left\{ \triangle \theta _{1}, \triangle \theta _{2},\cdots ,\triangle \theta _{n}\right\}$ for $n$ different domains, where each adapter is trained on its corresponding training splitted domain data $D_{i}$.
In the multi-domain lay paraphrasing scenario, $\forall x \in D_{multi}$, the generated text is expressed as:
\begin{equation}
y = \mathcal{M}(\theta, \alpha_{1} \cdot \triangle \theta_{1}, \cdots, \alpha_{n} \cdot \triangle \theta_{n}, x)
\end{equation}
where $\theta$ is the original parameters of the LLM $\mathcal{M}$, and $\alpha$ is the weight generated from the LoRA adapters weight generator.

\subsection{Sci-LoRA Framework}
\begin{figure*}[htp] 
 \center{\includegraphics[height=8cm,width=\textwidth]{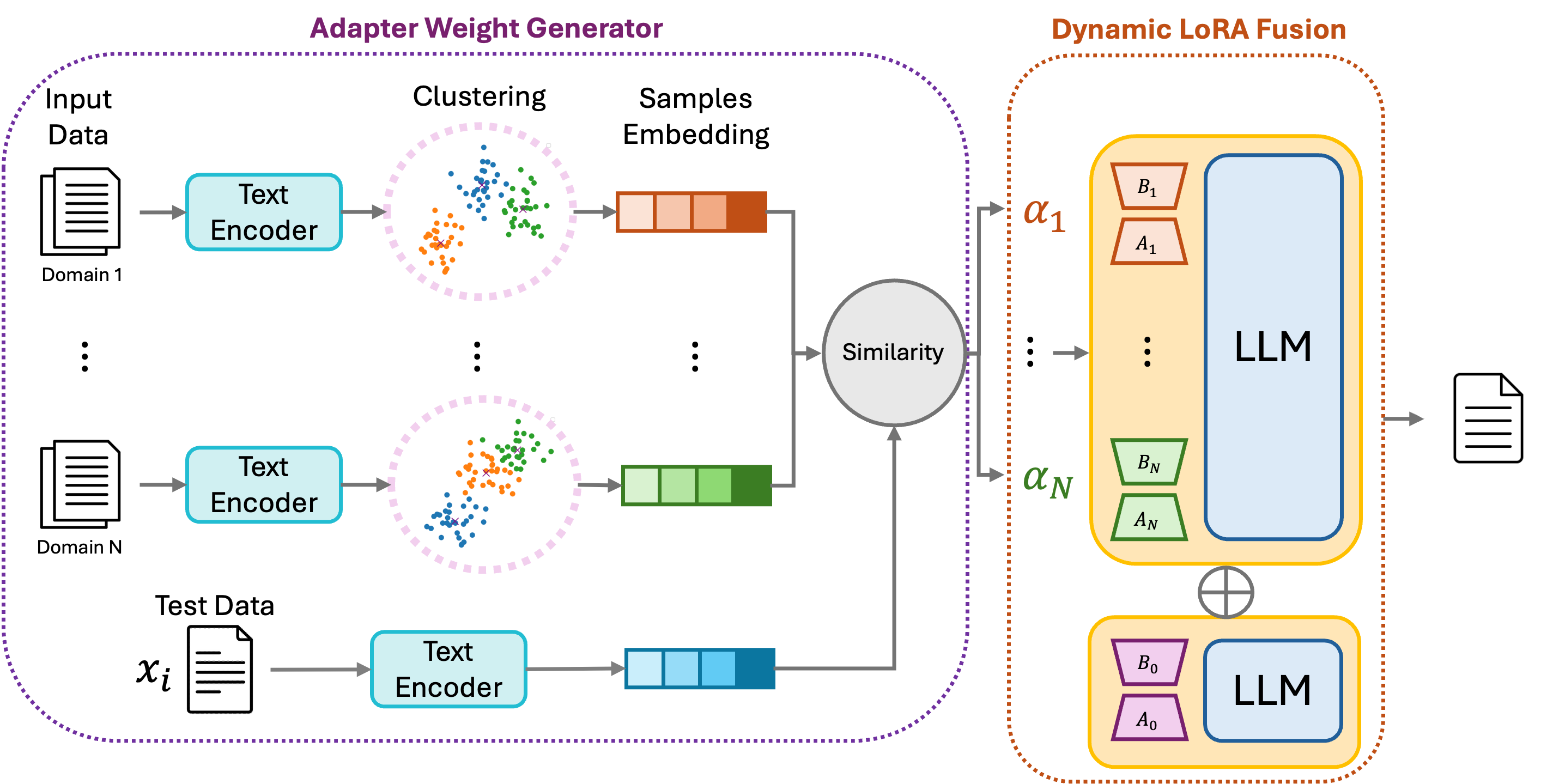}}
 \caption{\label{fig:framework} The Sci-LoRA framework including the adapter weight generator and dynamic LoRA fusion. All text encoder modules are the same text encoder trained on a subset of training data across all domains.}
 \end{figure*}
 
In this section, we describe our proposed Sci-LoRA as shown in Figure~\ref{fig:framework} for serving the mixture of LoRAs in multi-domain scenarios.
Our proposed model Sci-LoRA contains three major components: 
the domain LoRAs training module (Sec.~\ref{sec:training}), the adapter weight generator (Sec.~\ref{sec:AWG}), and the dynamic LoRA fusion strategy (Sec.~\ref{sec:fusion}).

\subsubsection{Domain LoRAs Training}~\label{sec:training}
We use the pre-trained Qwen2.5-7B-Instruct as the base model for Sci-LoRA because it is an open-source model under the Apache-2.0 license and it shows good performance for generating long texts~\cite{qwen2.5}. 
For each specific domain, we train one LoRA adapter on the domain training set based on Qwen2.5-7B-Instruct model. In total, we train twelve LoRA adapters corresponding to twelve different domains.
During the inference stage, Sci-LoRA can dynamically select the weighted mixture of LoRAs for a given text without domain tags. 

\subsubsection{Adapter Weight Generator}~\label{sec:AWG}
Although the input text originates from a specific domain, it often incorporates interdisciplinary technical knowledge.
To leverage the strengths of different domain-specific LoRAs, we design an Adapter Weight Generator that effectively and dynamically integrates and utilizes LoRAs across multiple domains. 
The Adapter Weight Generator includes a trained text encoder, and a weight generator.

\paragraph{Text Encoder} Directly using a pre-trained text encoder to obtain embeddings for the input text results in highly similar cosine similarity values with embeddings from various domain representations (Sec.~\ref{sec:ablation} and Sec.~\ref{sec:distance}).
To better differentiate representations among domains, we fine-tune a text encoder, Sentence-BERT~\cite{reimers-gurevych-2019-sentence}), using contrastive learning. The text encoder is fine-tuned on a subset of training data across all domains.
Specifically, the training dataset $D$ consists of positive paired samples $(x_{i}, x_{j})$ where $x_{j}:j \neq i, x_{j} \in Dom(x_{i})$ from the same domain, and negative paired samples $(x_{i}, x_{k})$ where $x_{k}: x_{k} \notin Dom(x_{k})$ randomly selected from other domains.
The training process is achieved through a contrastive loss~\cite{oord2018representation} as:
\begin{equation}
\mathcal{L} = \frac{e^{-\left\| x_{i}- x_{j}\right\|^{2}/\tau }}{e^{-\left\| x_{i}- x_{j}\right\|^{2}/\tau } + \sum_{k=1}^{m}e^{-\left\| x_{i}- x_{k}\right\|^{2}/\tau }}
\end{equation}
where $m$ is the number of negative pairs to each positive pair, and $\tau$ is the softmax temperature.

\paragraph{Weight Generator} Since each LoRA is fine-tuned on domain-specific textual data, it can be represented by a set of domain-related data points.
Rather than using the average embedding of randomly sampled data points from the training set, which can overlook important patterns in the data, we employ the k-means clustering algorithm to identify $k$ clusters within the embeddings of the training set. 
Random sampling may select over or under-representation of certain parts of the dataset while clusting-based approach ensures a more balanced and systematic coverage.
To ensure a more representative and structured sampling approach, we select the data points closest to the centroids of these clusters:
\begin{equation}
e_{j}^{*} = \arg\min_{E(x_{i}) \in C_{j}} \left\| E(x_{i}) - c_{j} \right\|_{2}
\end{equation}
where $c_{j} = \frac{1}{\left | C_{j}\right |}\sum_{E(x_{i}) \in C_{j}}^{}E(x_{i})$ is the centroid for its corresponding cluster $C_{j}$, and $E(\cdot)$ is the fine-tuned Text Encoder. 
The final representation for each domain-specific adapter is expressed as:
\begin{equation}~\label{equ:representation}
r_{\triangle \theta_{i}} = \frac{1}{K}\sum_{j=1}^{K}e_{j}^{*}
\end{equation}
where $K$ is the total number of clusters.

During inference, the weight $\alpha_{i}$ of input text $x_{i}$ for its domain adapter $\triangle \theta_{i}$ can be generated as:
\begin{equation}~\label{equ:weight}
\alpha _{i} = \frac{1}{1+\left\| E(x_{i})-r_{\triangle \theta_{i}}\right\|_{2}}
\end{equation}
where $E(\cdot)$ is the fine-tuned text encoder and $r_{\triangle \theta_{i}}$ is from Eq.~\ref{equ:representation}.

\begin{table*}[]
\caption{Dataset statistics reported in the format of technical
(source) / non-technical (target) summaries.}
\label{tab:dataset}
\centering
\begin{tabular}{lccccc}
\hline
Dataset  & \#Doc. & Avg. \# Sentence & Avg. Sentence Len. & FKGL          & DCRS          \\ \hline
PLOS     & 27,525 & 11.99 / 8.32     & 23.30 / 23.51      & 15.04 / 14.76 & 11.06 / 10.91 \\
eLife    & 4,828  & 7.18 / 17.97     & 23.14 / 20.56      & 15.57 / 10.92 & 11.78 / 8.83  \\
CELLS    & 62,886  & 12.82 / 7.54     & 22.08 / 22.63      & 16.79 / 16.67 & 6.78 / 7.01  \\
SciTechNews    & 2,431  & 8.78 / 6.59     & 22.31 / 23.87      & 16.33 / 15.80 & 7.14 / 7.59  \\
VTechAGP & 4,409  & 13.68 / 11.66    & 27.68 / 26.38      & 15.74 / 14.78 & 6.92 / 6.72   \\ \hline
\end{tabular}
\end{table*}
\subsubsection{Dynamic LoRA Fusion}~\label{sec:fusion}
To effectively balance domain-specific knowledge and generalization, we propose dynamic LoRA fusion module consisting of domain-specific LoRA aggregation and unified LoRA for multi-domain generalization.
First, all domain-specific LoRAs are merged with learned LoRA weights $\alpha$ in Eq.~\ref{equ:weight}.
We proceed to integrate these LoRA parameters $\triangle \theta$ into the LLM with parameter $\theta$. The specialized representation for text input $x_{i}$ is expressed as:
\begin{equation}
r_{specialized} = \mathcal{M} (\theta + \sum_{i=1}^{n}a_{i}\triangle \theta_{i}, x_{i})
\end{equation}
where $n$ is the number of domains.
Next, a single LoRA $\triangle \theta_{0}$ is trained on data across all domains to capture more generalized features. The generalized representation for input text $x_{i}$ is expressed as:
\begin{equation}
r_{generalized} = \mathcal{M}(\theta+\triangle \theta_{0}, x_{i})
\end{equation}
where $\theta$ is the original parameter for model $\mathcal{M}$.
This ensures that broad patterns across domains are captured, avoiding overfitting to specific domain.
Finally, we combine $r_{specialized}$ and $r_{generalized}$ as:
\begin{equation}
\hat{r} = \beta \cdot r_{specialized} + (1-\beta)
\cdot r_{generalized}
\end{equation}
where $\beta$ controls the balance between domain-specialized and general knowledge.

\section{Experiments}
\subsection{Evaluation Settings}
\subsubsection{Datasets}

In our experiments, we evaluate our proposed model over five public datasets: PLOS~\cite{goldsack-etal-2022-making}, eLife~\cite{goldsack-etal-2022-making}, CELLS~\cite{guo2024retrieval}, SciTechNews~\cite{cardenas-etal-2023-dont}, and VTechAGP~\cite{cheng-etal-2025-vtechagp}.
All these public datasets include technical abstracts and corresponding non-technical summaries.
We follow the same training, validation, and testing splits as these datasets originally have used~\footnote{We re-split SciTechNews as the original training set does not have the abstract and summary pairs.}.
Table~\ref{tab:dataset} provides detailed statistics of each dataset. 
Description of each dataset and explanation of Table~\ref{tab:dataset} are provided in Sec.~\ref{sec:data_appendix} in the Appendix.
Given the multiple domains and the unbalanced training data shown in Table~\ref{tab:dataset}, we conduct experiments to explore the generalization of Sci-LoRA and other LLMs for the cross-domain lay paraphrasing task.

\subsubsection{Baselines and Evaluation Metrics}
We compare Sci-LoRA with the following state-of-the-art LLM baselines: ChatGPT (gpt-3.5-turbo-0613)~\cite{NEURIPS2020_1457c0d6}, GPT-4o~\cite{achiam2023gpt}, Phi-3 (Phi-3-mini-128k-instruct)~\cite{abdin2024phi}, Phi-3.5 (Phi-3.5-mini-instruct), OPT (opt-6.7b)~\cite{zhang2022opt}, LLaMA3 (Llama-3.2-3B-Instruct)~\cite{dubey2024llama}, Qwen2.5 (Qwen2.5-7B-Instruct)~\cite{qwen2.5}, Mistral (Mistral-7B-Instruct-v0.3)~\cite{jiang2023mistral}, Mixtral (Mixtral-8x7B-Instruct-v0.1)~\cite{jiang2024mixtral}, and DSPT5~\cite{cheng-etal-2025-vtechagp}.
For evaluation metrics, we follow~\cite{cheng-etal-2025-vtechagp} to use embedding-based metrics: BERTScore (F1)~\cite{bert-score}, BLONDE (F1)~\cite{jiang-etal-2022-blonde}, sentence-level and document-level BLEU~\cite{lin-och-2004-orange}; Word-based metrics: ROUGE1, ROUGE2~\cite{lin-2004-rouge} and METEOR~\cite{banarjee2005}; End-to-end metrics: COMET~\cite{rei-etal-2022-searching}; Simplicity: SARI~\cite{xu-etal-2016-optimizing}, and Readability: FRES (Flesch Reading-Ease Score)~\cite{flesch1979write}.


\subsubsection{Parameter Settings}
\begin{table*}[]
\small
\centering
\caption{Results over six evaluation metrics. Another four metrics are reported in Table~\ref{tab:more_results}.}
\label{tab:main}
\begin{tabular}{l|cccccccccccc}
\hline
         & CELLS          & PLOS           & eLife          & News           & ALS            & AAD            & ENG            & LAHS           & NRE            & SCI            & BUS            & VM             \\ \hline
         & \multicolumn{12}{c}{d-BLEU (\%)}                                                                                                                                                                         \\ \hline
OPT      & 4.45           & 5.31           & 2.45           & 2.50           & 24.88          & 23.10          & 23.00          & 36.78          & 20.26          & 19.44          & 20.16          & 14.32          \\
LLaMA3   & 4.38           & 3.51           & 3.27           & 2.61           & 20.73          & 20.78          & 19.61          & 21.92          & 20.77          & 19.35          & 22.40          & 18.61          \\
Phi-4    & 6.31  & 6.30  & 1.90           & 1.32           & 25.55          & 28.71          & 24.45          & 36.59          & 24.42          & 20.96          & 29.67          & 21.78          \\
Mistral  & 5.27           & 4.47           & 3.12           & 2.61           & 28.62 & 28.63          & 24.53 & 22.14          & 11.55          & 17.18          & 16.32          & 13.66          \\
Mixtral  & 6.36           & 7.43           & 3.05           & 2.88           & 13.64          & 28.12          & 14.96          & 20.00          & 13.58          & 14.93          & 13.12          & 7.29           \\
Qwen2.5  & 9.26           & 10.18          &3.98  & 4.26           & 25.55          & 28.71          & 24.40          & 36.59          & 24.42          & 20.56          & 23.55          & 24.33          \\
GPT-3.5  & 6.70           & 7.08           & 1.74           & 3.07           & 15.39          & 15.59          & 13.75          & 16.99          & 14.11          & 12.84          & 17.20          & 13.62          \\
GPT-4o   & 5.10           & 5.76           & 2.65           & 2.92           & 11.05          & 12.40          & 9.65           & 13.60          & 10.78          & 9.06           & 9.54           & 9.10           \\
DSPT5    & -              & -              & -              & -              & 24.95          & 33.53          & 24.98          & 38.80          & 28.11          & 21.31          & \textbf{35.31} & 23.42 \\
Sci-LoRA & \textbf{11.15} & \textbf{12.43} & \textbf{6.09}  & \textbf{4.61}  & \textbf{31.03} & \textbf{38.97} & \textbf{28.31} & \textbf{40.33} & \textbf{29.61} & \textbf{23.31} & 32.86          & \textbf{29.55} \\ \hline
         & \multicolumn{12}{c}{BERTScore (F1 \%)}                                                                                                                                                                   \\ \hline
OPT      & 81.53          & 82.47          & 77.53          & 76.08          & 82.09          & 81.01          & 79.75          & 82.41          & 83.64          & 79.58          & 84.60          & 82.30          \\
LLaMA3   & 80.22          & 75.11          & 74.86          & 77.12          & 82.59          & 80.72          & 81.57          & 82.72          & 82.60          & 81.44          & 83.20          & 81.64          \\
Phi-4    & 82.64          & 83.04          & 78.28          & 76.20          & 84.98          & 82.55          & 83.57          & 86.24          & 84.75          & 82.91          & 85.22          & 84.37          \\
Mistral  & 79.36          & 80.29          & 80.48          & 77.01          & 84.69          & 81.86          & 83.31          & 82.82          & 81.27          & 80.13          & 80.73          & 81.92          \\
Mixtral  & 81.38          & 82.04          & 80.49          & 76.80          & 80.92          & 77.09          & 79.26          & 80.79          & 78.50          & 78.93          & 76.28          & 75.09          \\
Qwen2.5  & 82.36          & 82.70          & 80.72          & 77.90          & 84.98          & 82.55          & 83.16          & 86.24          & 84.75          & 82.85          & 84.11          & 84.37          \\
GPT-3.5  & 82.14          & 82.62          & 80.35          & 77.83          & 84.67          & 82.48          & 83.52          & 84.75          & 84.65          & 83.05          & 85.17          & 83.89          \\
GPT-4o   & 81.13          & 81.81          & 80.23          & 77.34          & 83.55          & 81.80          & 81.81          & 83.88          & 83.43          & 81.32          & 83.46          & 82.53          \\
DSPT5    & -              & -              & -              & -              & 85.48          & 83.70          & \textbf{84.41} & \textbf{87.27} & 85.51          & 82.90          & \textbf{87.97} & 84.20          \\
Sci-LoRA & \textbf{83.00} & \textbf{83.35} & \textbf{81.40} & \textbf{78.82} & \textbf{86.01} & \textbf{84.37} & 84.30          & 87.25          & \textbf{86.18} & \textbf{83.51} & 86.51          & \textbf{85.90} \\ \hline
         & \multicolumn{12}{c}{BLONDE (F1 \%)}                                                                                                                                                                      \\ \hline
OPT      & 14.08          & 5.52           & 4.08           & 4.95           & 33.08          & 38.01          & 9.05           & 43.37          & 30.89          & 8.35           & 35.21          & 11.03          \\
LLaMA3   & 14.42          & 5.46           & 4.38           & 4.45           & 31.90          & 42.79          & 9.92           & 45.22          & 31.51          & 8.43           & 43.49          & 11.43          \\
Phi-4    & 17.97          & 5.07           & 3.94           & 6.14           & 36.02          & 40.64          & 9.52           & 46.14          & 34.18          & 8.57           & 43.33          & 12.32          \\
Mistral  & 16.28          & 10.87          & 4.67           & 7.23           & \textbf{37.09} & 44.69          & 9.54           & 42.12          & 30.38          & 8.72           & 38.01          & 11.45          \\
Mixtral  & 17.15          & 19.73          & \textbf{5.07}  & 4.07           & 27.48          & 38.50          & 11.02          & 31.56          & 24.63          & 10.77          & 25.64          & 8.30           \\
Qwen2.5  & \textbf{18.80} & \textbf{21.46} & 4.97           & 9.10           & 36.02          & 40.64          & 9.27           & 46.14          & 34.18          & 8.47           & 34.22          & 31.32          \\
GPT-3.5  & 16.78          & 17.85          & 4.19           & 4.03           & 27.73          & 30.99          & \textbf{11.80} & 30.67          & 26.70          & 11.13          & 11.54          & 21.79          \\
GPT-4o   & 16.19          & 17.61          & 4.81           & 4.17           & 22.58          & 26.83          & 5.93           & 27.14          & 21.77          & 8.48           & 24.55          & 18.37          \\
DSPT5    & -              & -              & -              & -              & 36.75          & 44.11          & 9.62           & \textbf{48.51} & 36.60          & 8.22           & 35.53          & 30.69          \\
Sci-LoRA & 18.12          & 16.81          & 4.99           & \textbf{10.13} & 36.99          & \textbf{50.99} & 10.28          & 48.32          & \textbf{40.41} & \textbf{14.75} & \textbf{46.58} & \textbf{39.09} \\ \hline
         & \multicolumn{12}{c}{ROUGE1 (\%)}                                                                                                                                                                         \\ \hline
OPT      & 35.43          & 33.49          & 30.43          & 28.51          & 48.22          & 32.95          & 42.26          & 48.68          & 48.02          & 40.67          & 43.70          & 46.96          \\
LLaMA3   & 33.90          & 37.81          & 32.54          & 30.40          & 45.65          & 41.42          & 42.86          & 45.01          & 46.06          & 41.55          & 44.52          & 41.78          \\
Phi-4    & 39.46          & 40.17          & 28.11          & 25.17          & 52.90          & 46.51          & 48.39          & 56.03          & 51.67          & 46.05          & 52.17          & 48.60          \\
Mistral  & 31.44          & 33.42          & 35.63          & 28.09          & 52.35          & 45.33          & 48.20          & 52.32          & 47.74          & 43.41          & 43.36          & 47.94          \\
Mixtral  & 37.72          & 40.84          & 36.68          & 30.00          & 43.44          & 37.31          & 39.39          & 42.41          & 41.06          & 38.09          & 35.43          & 33.38          \\
Qwen2.5  & 40.55          & 42.75          & 37.57          & 31.85          & 52.90          & 46.51          & 47.54          & 56.03          & 51.67          & 45.58          & 47.53          & 48.60          \\
GPT-3.5  & 38.42          & 40.55          & 31.74          & 30.25          & 51.50          & 45.61          & 47.97          & 50.32          & 51.38          & 46.06          & 50.35          & 48.31          \\
GPT-4o   & 37.27          & 39.88          & 35.06          & 30.20          & 47.32          & 42.74          & 42.31          & 47.08          & 47.56          & 40.67          & 44.13          & 43.94          \\
DSPT5    & -              & -              & -              & -              & 54.48          & 50.84          & \textbf{51.73} & \textbf{59.67} & 56.47          & 47.14          & \textbf{60.75} & 50.28          \\
Sci-LoRA & \textbf{43.10} & \textbf{45.59} & \textbf{42.59} & \textbf{32.68} & \textbf{56.90} & \textbf{52.69} & 51.45          & 59.59          & \textbf{57.30} & \textbf{48.62} & 56.90          & \textbf{53.80} \\ \hline
         & \multicolumn{12}{c}{METEOR (\%)}                                                                                                                                                                         \\ \hline
OPT      & 27.70          & 25.45          & 21.70          & 20.47          & 37.92          & 28.90          & 34.35          & 44.62          & 33.88          & 32.53          & 34.70          & 36.19          \\
LLaMA3   & 29.29          & 23.55          & 20.18          & \textbf{25.82} & 41.90          & 33.67          & 32.02          & 45.51          & 33.49          & 30.77          & 31.87          & 30.11          \\
Phi-4    & \textbf{36.83} & \textbf{36.75} & 17.74          & 22.03          & 40.88          & 39.85          & 39.08          & 49.26          & 41.31          & 37.15          & 43.09          & 37.56          \\
Mistral  & 27.98          & 24.97          & 20.70          & 21.61          & 41.24          & 41.58          & 40.94          & 44.38          & 37.82          & 34.10          & 33.86          & 37.62          \\
Mixtral  & 30.92          & 31.91          & 19.56          & 22.79          & 30.24          & 30.54          & 29.97          & 33.98          & 30.42          & 28.86          & 26.28          & 24.35          \\
Qwen2.5  & 30.40          & 31.88          & 21.21          & 22.90          & 40.88          & 39.85          & 38.47          & 49.26          & 41.31          & 36.20          & 37.67          & 37.56          \\
GPT-3.5  & 28.55          & 28.15          & 16.07          & 21.16          & 38.94          & 36.86          & 37.47          & 41.68          & 39.71          & 36.06          & 40.56          & 36.09          \\
GPT-4o   & 30.81          & 31.38          & 18.88          & 24.01          & 33.77          & 33.32          & 31.73          & 36.68          & 35.18          & 30.23          & 32.23          & 31.66          \\
DSPT5    & -              & -              & -              & -              & 40.50          & 42.51          & 40.74          & 52.67          & 45.05          & 36.65          & \textbf{50.56} & 38.01          \\
Sci-LoRA & 32.29          & 35.02          & \textbf{28.98} & 23.87          & \textbf{45.51} & \textbf{46.71} & \textbf{42.73} & \textbf{53.37} & \textbf{47.59} & \textbf{40.18} & 48.36          & \textbf{44.00} \\ \hline
         & \multicolumn{12}{c}{SARI}                                                                                                                                                                                \\ \hline
OPT      & 37.27          & 37.54          & 37.27          & 39.69          & 35.38          & 31.92          & 37.24          & 36.49          & 34.66          & 36.97          & 40.68          & 37.49          \\
LLaMA3   & 40.13          & 37.61          & 42.47          & 40.70          & 35.93          & 34.45          & 35.61          & 33.17          & 35.10          & 37.28          & 37.62          & 36.23          \\
Phi-4    & 40.95          & 42.26          & 40.46          & 37.83          & 40.14          & 39.39          & 38.73          & 38.98          & 38.49          & 38.61          & 42.96          & 41.13          \\
Mistral  & \textbf{41.69} & 38.72          & 45.20          & 43.68          & 40.86          & 43.41          & 41.32          & 32.40          & 33.09          & 34.92          & 32.48          & 36.45          \\
Mixtral  & 40.10          & 39.99          & 42.95          & 39.43          & 35.28          & 35.53          & 34.37          & 32.07          & 33.77          & 35.71          & 35.46          & 33.36          \\
Qwen2.5  & 40.38          & \textbf{40.08} & 44.03          & 43.02          & 40.14          & 39.39          & 38.74          & 38.98          & 38.49          & 38.74          & 39.15          & 41.13          \\
GPT-3.5  & 40.12          & 39.60          & 42.66          & 39.34          & 37.83          & 35.41          & 36.90          & 34.50          & 36.61          & 38.53          & 40.07          & 38.07          \\
GPT-4o   & 39.93          & 39.60          & 43.04          & 39.48          & 35.99          & 33.65          & 34.96          & 32.56          & 34.9           & 36.73          & 36.25          & 36.11          \\
DSPT5    & -              & -              & -              & -              & 37.31          & 36.01          & 38.21          & 38.95          & 37.23          & 37.11          & \textbf{48.67} & 36.50          \\
Sci-LoRA & 41.15          & 40.07          & \textbf{47.64} & \textbf{43.76} & \textbf{41.32} & \textbf{44.26} & \textbf{41.64} & \textbf{42.88} & \textbf{41.77} & \textbf{40.60} & 45.08          & \textbf{42.45} \\ \hline
\end{tabular}
\end{table*}

\begin{table*}[]
\small
\centering
\caption{Results of the remaining four evaluation metrics over all baselines. The abbreviations stands for CELLS, PLOS, eLife, SciTechNews, Agriculture and Life Science, Architecture, Arts and Design, Engineering, Liberal Arts and Human Sciences, Natural Resources and Environment, Science, Business, and Veterinary Medicine.}
\label{tab:more_results}
\begin{tabular}{l|cccccccccccc}
\hline
         & CELLS          & PLOS           & eLife          & News           & ALS            & AAD            & ENG            & LAHS           & NRE            & SCI            & BUS            & VM             \\ \hline
         & \multicolumn{12}{c}{s-BLEU (\%)}                                                                                                                                                                          \\ \hline
LLaMA3   & 1.84           & 1.35           & 0.58           & 0.84           & 8.37           & 14.42          & 12.72          & 15.36          & 7.02           & 5.20           & 12.61          & 8.20           \\
Phi-4    & \textbf{4.08}  & \textbf{4.58}  & 0.44           & 1.33           & 8.95           & 13.43          & 9.72           & 19.82          & 7.55           & 8.22           & 16.03          & 9.32           \\
Mistral  & 1.51           & 1.42           & 0.85           & 0.76           & \textbf{11.46} & 14.70          & \textbf{13.74} & 11.72          & 5.51           & 3.89           & 12.10          & 10.37          \\
Mixtral  & 2.55           & 2.58           & 0.82           & 0.92           & 4.03           & 4.52           & 3.94           & 6.52           & 4.40           & 4.05           & 1.75           & 4.05           \\
Qwen2.5  & 3.07           & 3.23           & \textbf{1.22}  & 1.58           & 8.95           & 13.43          & 9.72           & 19.82          & 7.55           & 8.22           & 15.49          & 9.32           \\
GPT-3.5  & 2.56           & 2.58           & 0.75           & 1.10           & 4.53           & 5.74           & 4.34           & 6.25           & 4.08           & 4.20           & 8.17           & 4.51           \\
GPT-4o   & 2.15           & 2.12           & 0.75           & 0.88           & 3.84           & 3.61           & 3.51           & 6.70           & 4.06           & 3.32           & 3.42           & 3.55           \\
DSPT5    & -              & -              & -              & -              & 10.84          & 14.09          & 11.25          & 22.12          & 9.19           & 8.13           & \textbf{20.30} & \textbf{11.38} \\
Sci-LoRA & \textbf{3.99}  & \textbf{4.06}  & \textbf{1.22}  & \textbf{2.20}  & \textbf{10.21} & \textbf{18.38} & \textbf{11.69} & \textbf{23.99} & \textbf{13.84} & \textbf{9.86}  & 15.76          & \textbf{11.32} \\ \hline
         & \multicolumn{12}{c}{ROUGE2 (\%)}                                                                                                                                                                          \\ \hline
OPT      & 9.77           & 10.91          & 5.77           & 5.79           & 26.78          & 21.90          & 22.67          & 33.34          & 21.11          & 20.79          & 21.38          & 14.29          \\
LLaMA3   & 7.97           & 6.11           & 6.51           & 5.28           & 24.62          & 23.17          & 22.99          & 36.49          & 24.84          & 22.53          & 15.58          & 12.09          \\
Phi-4    & \textbf{14.59} & 15.80          & 5.86           & 4.72           & 27.50          & 23.37          & 23.38          & 35.59          & 26.53          & 21.45          & 29.32          & 22.02          \\
Mistral  & 8.98           & 6.75           & 7.76           & 4.07           & 28.99          & 28.18          & 26.58          & 30.50          & 23.37          & 18.78          & 18.36          & 12.98          \\
Mixtral  & 9.83           & 11.26          & 9.09           & 4.92           & 16.25          & 18.98          & 13.83          & 18.92          & 15.82          & 13.81          & 14.14          & 11.44          \\
Qwen2.5  & 11.92          & 13.25          & 9.23           & 6.23           & 27.50          & 23.37          & 23.20          & 35.59          & 26.53          & 20.59          & 23.35          & 22.02          \\
GPT-3.5  & 9.54           & 10.50          & 7.63           & 5.06           & 20.08          & 16.75          & 17.88          & 21.89          & 19.67          & 16.55          & 21.76          & 17.54          \\
GPT-4o   & 8.70           & 9.81           & 8.04           & 4.96           & 14.87          & 13.54          & 12.43          & 17.23          & 14.75          & 11.36          & 14.58          & 12.15          \\
DSPT5    & -              & -              & -              & -              & 28.58          & 27.55          & 27.02          & 40.24          & 30.83          & 21.97          & \textbf{39.61} & 24.63          \\
Sci-LoRA & 14.02          & \textbf{15.89} & \textbf{11.31} & \textbf{6.90}  & \textbf{32.16} & \textbf{30.98} & \textbf{27.62} & \textbf{40.92} & \textbf{33.19} & \textbf{24.26} & 34.50          & \textbf{28.24} \\ \hline
         & \multicolumn{12}{c}{COMET (\%)}                                                                                                                                                                           \\ \hline
OPT      & 74.74          & 77.73          & 74.74          & 71.86          & 80.45          & 71.81          & 81.42          & 77.90          & 78.89          & 76.13          & 81.01          & 76.41          \\
LLaMA3   & 79.17          & 77.43          & 78.40          & 74.80          & 82.92          & 76.06          & 82.80          & 81.87          & 84.54          & 80.07          & 83.39          & 83.44          \\
Phi-4    & 78.87          & 80.05          & 73.62          & 70.84          & 83.24          & 76.06          & 82.02          & 81.86          & 83.23          & 79.90          & 83.56          & 83.58          \\
Mistral  & 77.84          & 78.45          & 80.13          & 74.46          & 81.05          & 72.00          & 79.22          & 79.72          & 81.01          & 79.23          & 81.31          & 82.42          \\
Mixtral  & 79.46          & 79.26          & 77.98          & 74.24          & 78.57          & 68.67          & 75.52          & 75.98          & 73.99          & 74.46          & 69.64          & 68.16          \\
Qwen2.5  & 78.95          & 79.90          & 79.09          & 74.32          & 83.42          & 76.06          & 81.22          & 81.86          & 83.23          & 80.33          & 82.60          & 83.58          \\
GPT-3.5  & \textbf{80.23} & 79.69          & 77.20          & 75.22          & \textbf{85.49} & \textbf{80.30} & \textbf{84.00} & \textbf{83.29} & \textbf{85.79} & \textbf{82.98} & 85.29          & 84.10          \\
GPT-4o   & 79.97          & 80.26          & 77.50          & 75.39          & 84.47          & 78.71          & 81.72          & 82.33          & 84.95          & 80.67          & 83.86          & 83.71          \\
DSPT5    & -              & -              & -              & -              & 81.37          & 75.90          & 80.64          & 81.76          & 82.09          & 77.15          & 84.07          & 78.36          \\
Sci-LoRA & 79.10          & \textbf{80.29} & \textbf{82.95} & \textbf{76.00} & 83.79          & 78.02          & 82.90          & 82.62          & 84.66          & 80.87          & \textbf{85.40} & \textbf{84.67} \\ \hline
         & \multicolumn{12}{c}{FRES}                                                                                                                                                                                 \\ \hline
OPT      & 34.05          & 34.26          & 34.05          & 39.77          & 34.05          & 40.87          & 31.62          & 29.69          & 42.11          & 32.73          & 21.33          & 32.43          \\
LLaMA3   & 49.55          & 50.57          & 50.36          & \textbf{49.86} & 50.16          & 49.75          & 41.29          & 39.67          & 49.25          & 40.79          & 40.58          & 49.86          \\
Phi-4    & 34.15          & 33.54          & 23.26          & 33.44          & 35.07          & 51.89          & 32.63          & 30.50          & 42.82          & 33.65          & 21.43          & 42.31          \\
Mistral  & \textbf{51.06} & \textbf{53.10} & 53.61          & 42.72          & 33.75          & 50.57          & 31.72          & \textbf{41.09} & \textbf{51.38} & \textbf{43.63} & \textbf{41.70} & 43.02          \\
Mixtral  & 49.35          & 49.15          & 48.94          & 41.40          & 41.40          & 41.29          & 32.73          & 39.37          & 41.09          & 41.19          & 39.16          & 40.08          \\
Qwen2.5  & 43.43          & 43.43          & 53.41          & 42.11          & 35.07          & \textbf{51.98} & 32.43          & 30.50          & 42.82          & 33.34          & 23.04          & 42.31          \\
GPT-3.5  & 41.29          & 41.80          & 40.69          & 33.14          & 42.11          & 42.41          & 33.85          & 31.72          & 41.70          & 34.15          & 29.69          & 42.21          \\
GPT-4o   & 41.90          & 42.00          & 49.04          & 42.31          & \textbf{50.97} & 43.83          & \textbf{42.31} & 40.99          & 50.36          & 42.11          & 32.83          & \textbf{50.16} \\
DSPT5    & -              & -              & -              & -              & 33.65          & 41.50          & 31.62          & 28.17          & 41.19          & 32.33          & 21.74          & 33.34          \\
Sci-LoRA & 44.36          & 44.36          & \textbf{54.93} & 41.38          & 44.86          & 41.80          & 33.82          & 32.48          & 42.80          & 33.44          & 22.53          & 34.12          \\ \hline
\end{tabular}
\end{table*}

\begin{table*}[]
\small
\centering
\tabcolsep=0.13cm
\caption{Ablation study over Sci-LoRA components in College of Architecture, Arts, and Design domain from VTechAGP Dataset. Due to limited space, ablation results on all domains are in Table~\ref{tab:ablation_all} and Table~\ref{tab:ablation_all_2} in Appendix.}
\label{tab:ablation}
\begin{tabular}{lcccccccccc}
\hline
Model              & s-BLEU         & d-BLEU         & BERTScore      & BLONDE         & ROUGE1         & ROUGE2         & METEOR         & COMET          & SARI           & FRES           \\ \hline
Pre-trained        & 5.63           & 12.69          & 81.97          & 27.56          & 42.46          & 13.87          & 33.98          & \textbf{80.05} & 34.64          & 35.17          \\
Multi-LoRAs        & 19.78          & 34.39          & 83.30          & 46.43          & 47.83          & 26.51          & 41.77          & 77.20          & 40.76          & 42.41          \\
Single LoRA           & 13.43          & 28.71          & 82.55          & 40.64          & 46.51          & 23.37          & 39.85          & 76.06          & 39.39          & \textbf{51.89} \\
AWG\textsubscript{Random}     & 23.08          & 34.74          & 83.46          & 48.14          & 48.81          & 26.16          & 40.48          & 77.30          & 41.58          & 42.41          \\
AWG\textsubscript{K-Means}    & \textbf{23.17} & 37.08          & 83.72          & 50.02          & 51.13          & 28.28          & 43.91          & 77.83          & 43.56          & 42.31          \\
AWG\textsubscript{Contrastive}        & 18.06          & 38.62          & 84.03          & 50.00          & 52.07          & 30.86          & 44.89          & 77.97          & 44.00          & 41.29          \\
w/o Fusion & 14.86          & 31.22          & 81.59          & 44.02          & 50.82          & 22.62          & 41.42          & 78.59          & 43.63          & 44.03          \\ \hline
Sci-LoRA           & 18.38          & \textbf{38.97} & \textbf{84.37} & \textbf{50.99} & \textbf{52.69} & \textbf{30.98} & \textbf{46.71} & 78.02          & \textbf{44.26} & 41.80          \\ \hline
\end{tabular}
\end{table*}
We implement Sci-LoRA in PyTorch and finetune vairous LoRA adapters for each domain using LLaMA-Factory~\cite{zheng2024llamafactory}.
For domain LoRAs training, the learning rate is 1e-4, batch size is 4 per device, LoRA rank is 8, document maximum length is 2048.
For adapter weight generator, the learning rate is 1e-5, batch size is 16, sampling size is 500, the number of clusters is 10.
It is chosen by grid search  $\left\{5, 10, 15, 20 \right\}$ on the validation set.
The weight parameter $\beta$ is 0.5.
We use the same prompt "Generate another version of the provided text for general audiences" for Sci-LoRA and all baselines.
For LoRA tuning Sci-LoRA, we follow the early stopping strategy when selecting the model for testing. The model is evaluated on the validation set after every training epoch. The time for training is around 3 hours. The experiments are conducted on eight Nvidia A100 GPUs.

\subsection{Results and Analysis}
\subsubsection{Main Results}

The experimental results across five datasets spanning twelve domains are presented in Table~\ref{tab:main} and Table~\ref{tab:more_results}.
Note that Mixtral, GPT-3.5, and GPT-4o are pre-trained models, DSPT5 is a full-size model fine-tuned separately for each domain, and all other models utilize LoRA fine-tuning applied to a fusion of all data.
We observe that: (1) In general, fine-tuned models outperform pre-trained generalist models such as Mixtral and GPT-4o, highlighting the necessity of fine-tuning for specialized tasks. 
(2) The fully fine-tuned DSPT5 model outperforms other LoRA fine-tuned baselines because it is fine-tuned separately for each domain. 
Note that other LoRA refers to a single generalist LoRA finetuned on all domains.
This ensures that its performance is not influenced by data from other domains, allowing it to specialize more effectively. In contrast, the LoRA fine-tuned baselines use a single LoRA adapter trained on data from all domains, which may lead to cross-domain interference and reduced specialization. 
(3) Our proposed Sci-LoRA achieves the best performance among all pre-trained models, LoRA fine-tuned models, and separate domain fine-tuned DSPT5. We think that Sci-LoRA effectively mitigates cross-domain interference while improving generalization by dynamically fusing multiple LoRAs with adaptive weighting.

\subsubsection{Ablation Study}~\label{sec:ablation}

To assess the performance of each component in Sci-LoRA, we conducted the ablation study as shown in Table~\ref{tab:ablation}. Note that all models use the same base model of Qwen2.5-7B-Instruct for fair comparison.
We explore different methods of adapters training, adapters weight generation (AWG) and adapters weight fusion.
We observe that: (1) In general, fine-tuning separate LoRAs for each domain (Multi-LoRAs) yields better performance than using a single LoRA across all domains.
This is because a single LoRA captures more generalized information, making it less effective for domain-specific tasks.
However, Multi-LoRAs directly leverage fine-tuned domain adapters, assuming that domain information is available during inference.
(2) To fully leverage the benefits of fine-tuned domain-specific LoRAs, we explored various approaches for the adapter weight generator. Utilizing k-means for domain adapter representation learning outperforms random sampling followed by averaging embeddings for LoRA representation. Furthermore, incorporating a fine-tuned text encoder with contrastive learning further enhances domain adaptation performance.
(3) The dynamic fusion module plays a critical role in enhancing text generation quality. Without properly mixing LoRAs, embeddings from different adapters are simply averaged, which can negatively impact performance. By integrating all components, Sci-LoRA achieves the best results, demonstrating its effectiveness in cross-domain lay paraphrasing.

We also analyze the impact of fine-tuning a text encoder using contrastive learning by visualizing t-SNE embeddings from five datasets. Before fine-tuning, embeddings from different domains are intermingled, indicating that the pre-trained encoder does not effectively capture domain-specific features. After contrastive learning, domain separation improves, forming distinct clusters, though some overlap remains, particularly between CELLS and PLOS due to shared biomedical content. Additionally, some embeddings from different domains remain close, reflecting inherent semantic similarities, such as overlapping topics between SciTechNews and Life. More analysis is in Sec.~\ref{sec:distance}.

\subsubsection{Human Evaluation}
\begin{figure*}[htp]
    \centering
    \begin{subfigure} 
        \centering
        \includegraphics[height=6.5cm,width=7.75cm]{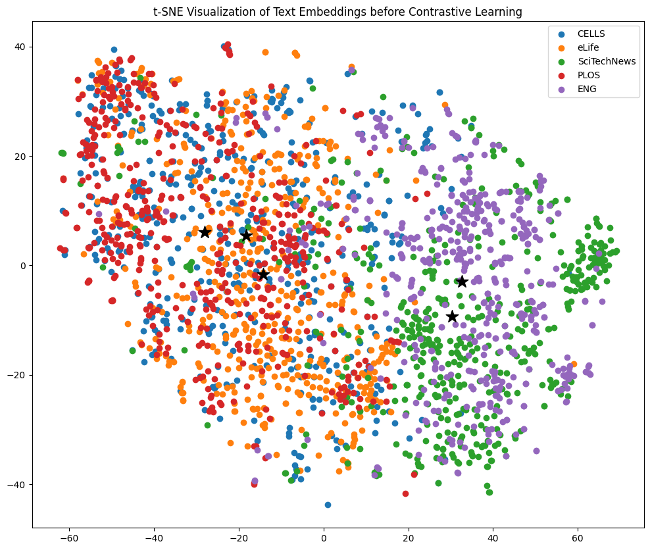} 
        \label{fig:subfigure1}
    \end{subfigure}
    \hfill 
    \begin{subfigure} 
        \centering
        \includegraphics[height=6.5cm,width=7.75cm]{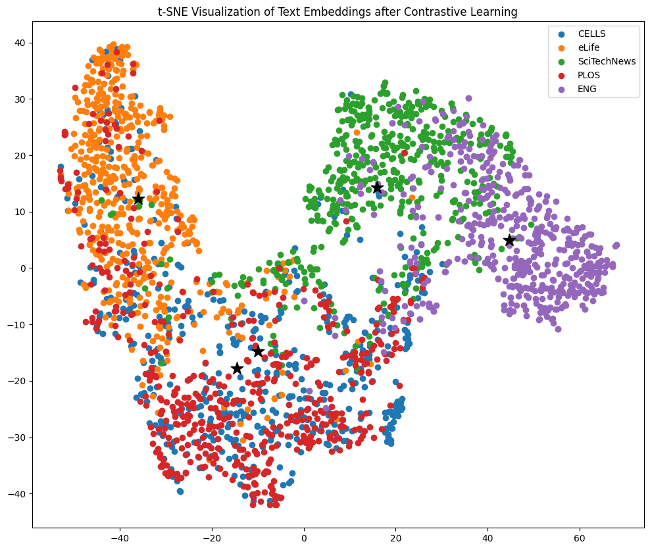} 
        \label{fig:subfigure2}
    \end{subfigure}
    \caption{t-SNE visualization (before contrastive learning and after contrastive learning) of text embeddings from CELLS, eLife, SciTechNews, PLOS and ENG in VTechAGP datasets.}
    \label{fig:visualization}
\end{figure*}

\begin{table}[]
\small
\centering
\tabcolsep=0.18cm
\caption{Mean human evaluation ratings 1-5 (the higher the better) of different models on all five datasets. Reported from left to right are: comprehensiveness, layness, meaning preservation, conciseness, and fluency.}
\label{tab:human}
\begin{tabular}{lcccccc}
\hline
         & COM           & LAY           & MP            & CON           & FLU    &ICC       \\ \hline
OPT      & 3.65          & 2.13          & 2.69          & 2.33          & 2.46  & 0.76        \\
LLaMA3   & 3.47          & \textbf{3.10} & 2.73          & 2.72          & 2.87   & 0.72       \\
Phi-4    & 3.53          & 2.77          & 2.27          & 2.20          & 2.71    & 0.75      \\
Mistral  & 3.48          & 2.64          & 2.73          & 2.93          & 2.80   & 0.28       \\
Mixtral  & 3.42          & 2.67          & 2.72          & 2.89          & 2.64   & 0.64       \\
Qwen2.5  & 3.78          & 2.86          & 2.80          & 2.77          & 3.31   & 0.75       \\
GPT-4o   & 3.25          & 2.68          & 3.08          & \textbf{3.53} & 3.15    & 0.75      \\
Sci-LoRA & \textbf{3.82} & 2.88          & \textbf{3.45} & 3.47          & \textbf{3.40}  & \textbf{0.80} \\ \hline
\end{tabular}
\end{table}

Following a similar setting as~\cite{cheng-etal-2025-vtechagp, liu-etal-2024-sumsurvey, li-etal-2024-side, song-etal-2024-finesure}, our evaluation uses a random sample of 15 abstracts from the total test split of all five datasets considering the workload. Judges are presented with both the academic abstract and generated general-audience abstracts from 8 models for each data sample in total of 120 abstracts. Using a 1-5 Likert scale, the judges are asked to rate the model output based on five criteria: comprehensiveness, layness, meaning preservation, conciseness, and fluency. 
The human evaluation setup is discussed in detail in Figure~\ref{fig:human} in the Appendix.

The human evaluation results are presented in Table~\ref{tab:human}. Overall, Sci-LoRA demonstrates the best performance in comprehensiveness, meaning preservation, and fluency.
Pre-trained models (i.e. Mixtral, GPT-4o) show high conciseness with minimal redundancy. 
However, they also miss detailed domain-specific contents due to their more generalized training paradigms. 
Fine-tuned models (i.e. OPT, Phi-4, etc.) can provide more comprehensive domain information but tend to produce verbose or repetitive outputs at the same time.
By dynamically selecting the most relevant domain knowledge, Sci-LoRA effectively integrates domain-specific information while maintaining the highest level of meaning preservation. It also avoids generating verbose explanations and unnecessary elaborations.
Fluency is primarily influenced by pre-training, as it is a general text generation metric. Since Sci-LoRA is built on Qwen2.5, it inherits high fluency from Qwen2.5.
For layness, all models have some room for improvement.  This challenge is particularly evident for highly specialized texts (i.e. biomedical domain text), where fully paraphrasing technical content into non-technical language without compromising meaning preservation remains difficult.
We also report the intraclass correlation coefficient (ICC) – average fixed raters ICC, which is used to determine if items or subjects can be rated reliably by different raters. We observe that GPT4o, Qwen2.5, Phi-4, OPT and ours Sci-LoRA show good reliability. Mixtral and LLaMA3 show moderate reliability. Mistral shows poor reliability.

\section{Conclusion}
In this paper, we propose Sci-LoRA~\footnote{Code: \href{https://github.com/gjiaying/Sci-LoRA}{https://github.com/gjiaying/Sci-LoRA}}, a mixture of scientific LoRAs designed for cross-domain lay paraphrasing.
Without requiring domain-specific information during inference, Sci-LoRA dynamically generates weights for domain-specific adapters. To enhance domain representation learning, a customized text encoder is fine-tuned using contrastive learning. Sci-LoRA then integrates these domain-specific adapters through a dynamic LoRA fusion module to facilitate cross-domain text generation.
We evaluate Sci-LoRA against multiple state-of-the-art baselines across 12 different domains and 5 distinct datasets, using 10 automatic evaluation metrics as well as human assessment. Extensive experimental results demonstrate that Sci-LoRA consistently outperforms existing SOTA models in the cross-domain lay paraphrasing task.

\section{Limitations}
Sci-LoRA has the following limitations currently: 
(1) Scaling Sci-LoRA to accommodate hundreds of different domains is difficult. The current implementation relies on PEFT~\footnote{https://huggingface.co/docs/peft/en/index} to load and merge LoRAs, with domain-specific LoRA weights computed dynamically during inference for each batch of input text. As the number of domains increases, inference latency is expected to grow significantly.
(2) Sci-LoRA does not support unseen domains without training data, as each domain-specific adapter requires fine-tuning on its respective domain data. To address this limitation, we will explore few-shot learning techniques for low-resource domains for the future work.
(3) Sci-LoRA is designed to integrate multiple LoRAs into a single base model. Our experiments have focused on Qwen2.5-7B-Instruct, as this pre-trained model has demonstrated strong inference performance. However, results may vary for different base models. We will investigate LoRA fusion across multiple pre-trained models to assess broader applicability in future work.

\bibliography{custom}

\begin{thebibliography}{59}
\providecommand{\natexlab}[1]{#1}

\bibitem[{Abdin et~al.(2024)Abdin, Aneja, Awadalla, Awadallah, Awan, Bach, Bahree, Bakhtiari, Bao, Behl et~al.}]{abdin2024phi}
Marah Abdin, Jyoti Aneja, Hany Awadalla, Ahmed Awadallah, Ammar~Ahmad Awan, Nguyen Bach, Amit Bahree, Arash Bakhtiari, Jianmin Bao, Harkirat Behl, et~al. 2024.
\newblock Phi-3 technical report: A highly capable language model locally on your phone.
\newblock \emph{arXiv preprint arXiv:2404.14219}.

\bibitem[{Achiam et~al.(2023)Achiam, Adler, Agarwal, Ahmad, Akkaya, Aleman, Almeida, Altenschmidt, Altman, Anadkat et~al.}]{achiam2023gpt}
Josh Achiam, Steven Adler, Sandhini Agarwal, Lama Ahmad, Ilge Akkaya, Florencia~Leoni Aleman, Diogo Almeida, Janko Altenschmidt, Sam Altman, Shyamal Anadkat, et~al. 2023.
\newblock Gpt-4 technical report.
\newblock \emph{arXiv preprint arXiv:2303.08774}.

\bibitem[{Al-Thanyyan and Azmi(2021)}]{al2021automated}
Suha~S Al-Thanyyan and Aqil~M Azmi. 2021.
\newblock Automated text simplification: a survey.
\newblock \emph{ACM Computing Surveys (CSUR)}, 54(2):1--36.

\bibitem[{Attal et~al.(2023)Attal, Ondov, and Demner-Fushman}]{attal2023dataset}
Kush Attal, Brian Ondov, and Dina Demner-Fushman. 2023.
\newblock A dataset for plain language adaptation of biomedical abstracts.
\newblock \emph{Scientific Data}, 10(1):8.

\bibitem[{Banerjee and Lavie(2005)}]{banarjee2005}
Satanjeev Banerjee and Alon Lavie. 2005.
\newblock {{METEOR}}: {{An}} automatic metric for {{MT}} evaluation with improved correlation with human judgments.
\newblock In \emph{Proceedings of the {{ACL}} Workshop on Intrinsic and Extrinsic Evaluation Measures for Machine Translation and/or Summarization}, pages 65--72, {Ann Arbor, Michigan}. {Association for Computational Linguistics}.

\bibitem[{Brown et~al.(2020)Brown, Mann, Ryder, Subbiah, Kaplan, Dhariwal, Neelakantan, Shyam, Sastry, Askell, Agarwal, Herbert-Voss, Krueger, Henighan, Child, Ramesh, Ziegler, Wu, Winter, Hesse, Chen, Sigler, Litwin, Gray, Chess, Clark, Berner, McCandlish, Radford, Sutskever, and Amodei}]{NEURIPS2020_1457c0d6}
Tom Brown, Benjamin Mann, Nick Ryder, Melanie Subbiah, Jared~D Kaplan, Prafulla Dhariwal, Arvind Neelakantan, Pranav Shyam, Girish Sastry, Amanda Askell, Sandhini Agarwal, Ariel Herbert-Voss, Gretchen Krueger, Tom Henighan, Rewon Child, Aditya Ramesh, Daniel Ziegler, Jeffrey Wu, Clemens Winter, Chris Hesse, Mark Chen, Eric Sigler, Mateusz Litwin, Scott Gray, Benjamin Chess, Jack Clark, Christopher Berner, Sam McCandlish, Alec Radford, Ilya Sutskever, and Dario Amodei. 2020.
\newblock \href {https://proceedings.neurips.cc/paper_files/paper/2020/file/1457c0d6bfcb4967418bfb8ac142f64a-Paper.pdf} {Language models are few-shot learners}.
\newblock In \emph{Advances in Neural Information Processing Systems}, volume~33, pages 1877--1901. Curran Associates, Inc.

\bibitem[{Buehler and Buehler(2024)}]{10.1063/5.0203126}
Eric~L. Buehler and Markus~J. Buehler. 2024.
\newblock \href {https://doi.org/10.1063/5.0203126} {X-lora: Mixture of low-rank adapter experts, a flexible framework for large language models with applications in protein mechanics and molecular design}.
\newblock \emph{APL Machine Learning}, 2(2):026119.

\bibitem[{Cai et~al.(2024)Cai, Jiang, Wang, Tang, Kim, and Huang}]{cai2024survey}
Weilin Cai, Juyong Jiang, Fan Wang, Jing Tang, Sunghun Kim, and Jiayi Huang. 2024.
\newblock A survey on mixture of experts.
\newblock \emph{arXiv preprint arXiv:2407.06204}.

\bibitem[{Cardenas et~al.(2023)Cardenas, Yao, Wang, and Hou}]{cardenas-etal-2023-dont}
Ronald Cardenas, Bingsheng Yao, Dakuo Wang, and Yufang Hou. 2023.
\newblock \href {https://doi.org/10.18653/v1/2023.emnlp-main.76} {{`}don{'}t get too technical with me{'}: A discourse structure-based framework for automatic science journalism}.
\newblock In \emph{Proceedings of the 2023 Conference on Empirical Methods in Natural Language Processing}, pages 1186--1202, Singapore. Association for Computational Linguistics.

\bibitem[{Cheng et~al.(2025)Cheng, Gong, Yuan, Ingram, Fox, and Eldardiry}]{cheng-etal-2025-vtechagp}
Ming Cheng, Jiaying Gong, Chenhan Yuan, William~A Ingram, Edward Fox, and Hoda Eldardiry. 2025.
\newblock \href {https://aclanthology.org/2025.naacl-long.311/} {{VT}ech{AGP}: An academic-to-general-audience text paraphrase dataset and benchmark models}.
\newblock In \emph{Proceedings of the 2025 Conference of the Nations of the Americas Chapter of the Association for Computational Linguistics: Human Language Technologies (Volume 1: Long Papers)}, pages 6110--6130, Albuquerque, New Mexico. Association for Computational Linguistics.

\bibitem[{Dubey et~al.(2024)Dubey, Jauhri, Pandey, Kadian, Al-Dahle, Letman, Mathur, Schelten, Yang, Fan et~al.}]{dubey2024llama}
Abhimanyu Dubey, Abhinav Jauhri, Abhinav Pandey, Abhishek Kadian, Ahmad Al-Dahle, Aiesha Letman, Akhil Mathur, Alan Schelten, Amy Yang, Angela Fan, et~al. 2024.
\newblock The llama 3 herd of models.
\newblock \emph{arXiv preprint arXiv:2407.21783}.

\bibitem[{Feng et~al.(2024)Feng, Hao, Zhang, Han, and Wang}]{feng-etal-2024-mixture}
Wenfeng Feng, Chuzhan Hao, Yuewei Zhang, Yu~Han, and Hao Wang. 2024.
\newblock \href {https://aclanthology.org/2024.lrec-main.994} {Mixture-of-{L}o{RA}s: An efficient multitask tuning method for large language models}.
\newblock In \emph{Proceedings of the 2024 Joint International Conference on Computational Linguistics, Language Resources and Evaluation (LREC-COLING 2024)}, pages 11371--11380, Torino, Italia. ELRA and ICCL.

\bibitem[{Flesch(1979)}]{flesch1979write}
Rudolf Flesch. 1979.
\newblock How to write plain {{English}}.
\newblock \emph{University of Canterbury. Available at http://www. mang. canterbury. ac. nz/writing\_guide/writing/flesch. shtml.[Retrieved 5 February 2016]}.

\bibitem[{Fonseca and Cohen(2024)}]{fonseca-cohen-2024-large-language}
Marcio Fonseca and Shay Cohen. 2024.
\newblock \href {https://doi.org/10.18653/v1/2024.findings-acl.508} {Can large language model summarizers adapt to diverse scientific communication goals?}
\newblock In \emph{Findings of the Association for Computational Linguistics: ACL 2024}, pages 8599--8618, Bangkok, Thailand. Association for Computational Linguistics.

\bibitem[{Giannouris et~al.(2024)Giannouris, Myridis, Passali, and Tsoumakas}]{giannouris-etal-2024-plain}
Polydoros Giannouris, Theodoros Myridis, Tatiana Passali, and Grigorios Tsoumakas. 2024.
\newblock \href {https://aclanthology.org/2024.determit-1.6} {Plain language summarization of clinical trials}.
\newblock In \emph{Proceedings of the Workshop on DeTermIt! Evaluating Text Difficulty in a Multilingual Context @ LREC-COLING 2024}, pages 60--67, Torino, Italia. ELRA and ICCL.

\bibitem[{Goldsack et~al.(2024)Goldsack, Scarton, Shardlow, and Lin}]{goldsack-etal-2024-overview}
Tomas Goldsack, Carolina Scarton, Matthew Shardlow, and Chenghua Lin. 2024.
\newblock \href {https://doi.org/10.18653/v1/2024.bionlp-1.10} {Overview of the {B}io{L}ay{S}umm 2024 shared task on the lay summarization of biomedical research articles}.
\newblock In \emph{Proceedings of the 23rd Workshop on Biomedical Natural Language Processing}, pages 122--131, Bangkok, Thailand. Association for Computational Linguistics.

\bibitem[{Goldsack et~al.(2022)Goldsack, Zhang, Lin, and Scarton}]{goldsack-etal-2022-making}
Tomas Goldsack, Zhihao Zhang, Chenghua Lin, and Carolina Scarton. 2022.
\newblock \href {https://doi.org/10.18653/v1/2022.emnlp-main.724} {Making science simple: Corpora for the lay summarisation of scientific literature}.
\newblock In \emph{Proceedings of the 2022 Conference on Empirical Methods in Natural Language Processing}, pages 10589--10604, Abu Dhabi, United Arab Emirates. Association for Computational Linguistics.

\bibitem[{Guo et~al.(2018)Guo, Shah, and Barzilay}]{guo-etal-2018-multi}
Jiang Guo, Darsh Shah, and Regina Barzilay. 2018.
\newblock \href {https://doi.org/10.18653/v1/D18-1498} {Multi-source domain adaptation with mixture of experts}.
\newblock In \emph{Proceedings of the 2018 Conference on Empirical Methods in Natural Language Processing}, pages 4694--4703, Brussels, Belgium. Association for Computational Linguistics.

\bibitem[{Guo et~al.(2024)Guo, Qiu, Leroy, Wang, and Cohen}]{guo2024retrieval}
Yue Guo, Wei Qiu, Gondy Leroy, Sheng Wang, and Trevor Cohen. 2024.
\newblock Retrieval augmentation of large language models for lay language generation.
\newblock \emph{Journal of Biomedical Informatics}, 149:104580.

\bibitem[{Guo et~al.(2021)Guo, Qiu, Wang, and Cohen}]{guo2021automated}
Yue Guo, Wei Qiu, Yizhong Wang, and Trevor Cohen. 2021.
\newblock Automated lay language summarization of biomedical scientific reviews.
\newblock In \emph{Proceedings of the AAAI Conference on Artificial Intelligence}, volume~35, pages 160--168.

\bibitem[{Han et~al.(2024)Han, Gao, Liu, Zhang, and Zhang}]{han2024parameter}
Zeyu Han, Chao Gao, Jinyang Liu, Jeff Zhang, and Sai~Qian Zhang. 2024.
\newblock Parameter-efficient fine-tuning for large models: A comprehensive survey.
\newblock \emph{arXiv preprint arXiv:2403.14608}.

\bibitem[{Houlsby et~al.(2019)Houlsby, Giurgiu, Jastrzebski, Morrone, De~Laroussilhe, Gesmundo, Attariyan, and Gelly}]{houlsby2019parameter}
Neil Houlsby, Andrei Giurgiu, Stanislaw Jastrzebski, Bruna Morrone, Quentin De~Laroussilhe, Andrea Gesmundo, Mona Attariyan, and Sylvain Gelly. 2019.
\newblock Parameter-efficient transfer learning for nlp.
\newblock In \emph{International conference on machine learning}, pages 2790--2799. PMLR.

\bibitem[{Hu et~al.(2022)Hu, Shen, Wallis, Allen-Zhu, Li, Wang, Wang, and Chen}]{hu2022lora}
Edward~J Hu, Yelong Shen, Phillip Wallis, Zeyuan Allen-Zhu, Yuanzhi Li, Shean Wang, Lu~Wang, and Weizhu Chen. 2022.
\newblock \href {https://openreview.net/forum?id=nZeVKeeFYf9} {Lo{RA}: Low-rank adaptation of large language models}.
\newblock In \emph{International Conference on Learning Representations}.

\bibitem[{Huang et~al.(2023)Huang, Liu, Lin, Pang, Du, and Lin}]{huang2023lorahub}
Chengsong Huang, Qian Liu, Bill~Yuchen Lin, Tianyu Pang, Chao Du, and Min Lin. 2023.
\newblock Lorahub: Efficient cross-task generalization via dynamic lora composition.
\newblock \emph{arXiv preprint arXiv:2307.13269}.

\bibitem[{Jiang et~al.(2023)Jiang, Sablayrolles, Mensch, Bamford, Chaplot, Casas, Bressand, Lengyel, Lample, Saulnier et~al.}]{jiang2023mistral}
Albert~Q Jiang, Alexandre Sablayrolles, Arthur Mensch, Chris Bamford, Devendra~Singh Chaplot, Diego de~las Casas, Florian Bressand, Gianna Lengyel, Guillaume Lample, Lucile Saulnier, et~al. 2023.
\newblock Mistral 7b.
\newblock \emph{arXiv preprint arXiv:2310.06825}.

\bibitem[{Jiang et~al.(2024)Jiang, Sablayrolles, Roux, Mensch, Savary, Bamford, Chaplot, Casas, Hanna, Bressand et~al.}]{jiang2024mixtral}
Albert~Q Jiang, Alexandre Sablayrolles, Antoine Roux, Arthur Mensch, Blanche Savary, Chris Bamford, Devendra~Singh Chaplot, Diego de~las Casas, Emma~Bou Hanna, Florian Bressand, et~al. 2024.
\newblock Mixtral of experts.
\newblock \emph{arXiv preprint arXiv:2401.04088}.

\bibitem[{Jiang et~al.(2022)Jiang, Liu, Ma, Zhang, Yang, Huang, Sennrich, Cotterell, Sachan, and Zhou}]{jiang-etal-2022-blonde}
Yuchen Jiang, Tianyu Liu, Shuming Ma, Dongdong Zhang, Jian Yang, Haoyang Huang, Rico Sennrich, Ryan Cotterell, Mrinmaya Sachan, and Ming Zhou. 2022.
\newblock \href {https://doi.org/10.18653/v1/2022.naacl-main.111} {{{BlonDe}}: {{An}} automatic evaluation metric for document-level machine translation}.
\newblock In \emph{Proceedings of the 2022 Conference of the North American Chapter of the Association for Computational Linguistics: {{Human}} Language Technologies}, pages 1550--1565, {Seattle, United States}. {Association for Computational Linguistics}.

\bibitem[{Kim et~al.(2024)Kim, Kanakarajan, and Sankarasubbu}]{kim2024saama}
Hwanmun Kim, Kamal~Raj Kanakarajan, and Malaikannan Sankarasubbu. 2024.
\newblock Saama technologies at biolaysumm: Abstract based fine-tuned models with lora.
\newblock In \emph{Proceedings of the 23rd Workshop on Biomedical Natural Language Processing}, pages 786--792.

\bibitem[{Li et~al.(2024)Li, Wu, Schlegel, Batista-Navarro, Madusanka, Zahid, Zeng, Wang, He, Li, and Nenadic}]{li-etal-2024-side}
Hao Li, Yuping Wu, Viktor Schlegel, Riza Batista-Navarro, Tharindu Madusanka, Iqra Zahid, Jiayan Zeng, Xiaochi Wang, Xinran He, Yizhi Li, and Goran Nenadic. 2024.
\newblock \href {https://doi.org/10.18653/v1/2024.findings-acl.9} {Which side are you on? a multi-task dataset for end-to-end argument summarisation and evaluation}.
\newblock In \emph{Findings of the Association for Computational Linguistics ACL 2024}, pages 133--150, Bangkok, Thailand and virtual meeting. Association for Computational Linguistics.

\bibitem[{Lin(2004)}]{lin-2004-rouge}
Chin-Yew Lin. 2004.
\newblock {{ROUGE}}: {{A}} package for automatic evaluation of summaries.
\newblock In \emph{Text Summarization Branches Out}, pages 74--81, {Barcelona, Spain}. {Association for Computational Linguistics}.

\bibitem[{Lin and Och(2004)}]{lin-och-2004-orange}
Chin-Yew Lin and Franz~Josef Och. 2004.
\newblock {{ORANGE}}: A method for evaluating automatic evaluation metrics for machine translation.
\newblock In \emph{{{COLING}} 2004: {{Proceedings}} of the 20th International Conference on Computational Linguistics}, pages 501--507, {Geneva, Switzerland}. {COLING}.

\bibitem[{Liu et~al.(2023)Liu, Wang, and Demberg}]{pu-etal-2023-incorporating}
Dongqi Liu, Yifan Wang, and Vera Demberg. 2023.
\newblock \href {https://doi.org/10.18653/v1/2023.acl-long.306} {Incorporating distributions of discourse structure for long document abstractive summarization}.
\newblock In \emph{Proceedings of the 61st Annual Meeting of the Association for Computational Linguistics (Volume 1: Long Papers)}, pages 5574--5590, Toronto, Canada. Association for Computational Linguistics.

\bibitem[{Liu et~al.(2024{\natexlab{a}})Liu, Wu, Liu, and Duan}]{liu2024learning}
Jialin Liu, Jianhua Wu, Jie Liu, and Yutai Duan. 2024{\natexlab{a}}.
\newblock Learning attentional mixture of loras for language model continual learning.
\newblock \emph{arXiv preprint arXiv:2409.19611}.

\bibitem[{Liu et~al.(2024{\natexlab{b}})Liu, Liu, Yu, Zhang, Jiang, Li, and Huang}]{liu-etal-2024-sumsurvey}
Ran Liu, Ming Liu, Min Yu, He~Zhang, Jianguo Jiang, Gang Li, and Weiqing Huang. 2024{\natexlab{b}}.
\newblock \href {https://doi.org/10.18653/v1/2024.findings-acl.574} {{S}um{S}urvey: An abstractive dataset of scientific survey papers for long document summarization}.
\newblock In \emph{Findings of the Association for Computational Linguistics ACL 2024}, pages 9632--9651, Bangkok, Thailand and virtual meeting. Association for Computational Linguistics.

\bibitem[{Liu et~al.(2024{\natexlab{c}})Liu, Feng, Han, Balachandran, Park, Kumar, and Tsvetkov}]{liu-etal-2024-p3sum}
Yuhan Liu, Shangbin Feng, Xiaochuang Han, Vidhisha Balachandran, Chan~Young Park, Sachin Kumar, and Yulia Tsvetkov. 2024{\natexlab{c}}.
\newblock \href {https://doi.org/10.18653/v1/2024.naacl-long.119} {{P}$^3${S}um: Preserving author{'}s perspective in news summarization with diffusion language models}.
\newblock In \emph{Proceedings of the 2024 Conference of the North American Chapter of the Association for Computational Linguistics: Human Language Technologies (Volume 1: Long Papers)}, pages 2154--2173, Mexico City, Mexico. Association for Computational Linguistics.

\bibitem[{Luo et~al.(2024)Luo, Lei, Lei, Liu, He, Zhao, and Liu}]{luo2024moelora}
Tongxu Luo, Jiahe Lei, Fangyu Lei, Weihao Liu, Shizhu He, Jun Zhao, and Kang Liu. 2024.
\newblock Moelora: Contrastive learning guided mixture of experts on parameter-efficient fine-tuning for large language models.
\newblock \emph{arXiv preprint arXiv:2402.12851}.

\bibitem[{Lv et~al.(2024)Lv, Li, Zhang, Chen, Qi, Zhang, and Zheng}]{lv-etal-2024-hyperlora}
Chuancheng Lv, Lei Li, Shitou Zhang, Gang Chen, Fanchao Qi, Ningyu Zhang, and Hai-Tao Zheng. 2024.
\newblock \href {https://doi.org/10.18653/v1/2024.findings-emnlp.956} {{H}yper{L}o{RA}: Efficient cross-task generalization via constrained low-rank adapters generation}.
\newblock In \emph{Findings of the Association for Computational Linguistics: EMNLP 2024}, pages 16376--16393, Miami, Florida, USA. Association for Computational Linguistics.

\bibitem[{Malik et~al.(2024)Malik, Pradeep, and Seth}]{malik-etal-2024-hgp}
Hemang Malik, Gaurav Pradeep, and Pratinav Seth. 2024.
\newblock \href {https://doi.org/10.18653/v1/2024.bionlp-1.78} {{HGP}-{NLP} at {B}io{L}ay{S}umm: Leveraging {L}o{RA} for lay summarization of biomedical research articles using {S}eq2{S}eq transformers}.
\newblock In \emph{Proceedings of the 23rd Workshop on Biomedical Natural Language Processing}, pages 831--836, Bangkok, Thailand. Association for Computational Linguistics.

\bibitem[{Muqeeth et~al.(2024)Muqeeth, Liu, Liu, and Raffel}]{muqeeth2024learning}
Mohammed Muqeeth, Haokun Liu, Yufan Liu, and Colin Raffel. 2024.
\newblock Learning to route among specialized experts for zero-shot generalization.
\newblock \emph{arXiv preprint arXiv:2402.05859}.

\bibitem[{Oord et~al.(2018)Oord, Li, and Vinyals}]{oord2018representation}
Aaron van~den Oord, Yazhe Li, and Oriol Vinyals. 2018.
\newblock Representation learning with contrastive predictive coding.
\newblock \emph{arXiv preprint arXiv:1807.03748}.

\bibitem[{Prabhakar et~al.(2024)Prabhakar, Li, Narasimhan, Kakade, Malach, and Jelassi}]{prabhakar2024lora}
Akshara Prabhakar, Yuanzhi Li, Karthik Narasimhan, Sham Kakade, Eran Malach, and Samy Jelassi. 2024.
\newblock Lora soups: Merging loras for practical skill composition tasks.
\newblock \emph{arXiv preprint arXiv:2410.13025}.

\bibitem[{Rei et~al.(2022)Rei, Farinha, {de Souza}, Ramos, Martins, Coheur, and Lavie}]{rei-etal-2022-searching}
Ricardo Rei, Ana~C Farinha, Jos{\'e}~G.C. {de Souza}, Pedro~G. Ramos, Andr{\'e}~F.T. Martins, Luisa Coheur, and Alon Lavie. 2022.
\newblock Searching for {{COMETINHO}}: {{The}} little metric that could.
\newblock In \emph{Proceedings of the 23rd Annual Conference of the European Association for Machine Translation}, pages 61--70, {Ghent, Belgium}. {European Association for Machine Translation}.

\bibitem[{Reimers and Gurevych(2019)}]{reimers-gurevych-2019-sentence}
Nils Reimers and Iryna Gurevych. 2019.
\newblock \href {https://doi.org/10.18653/v1/D19-1410} {Sentence-{BERT}: Sentence embeddings using {S}iamese {BERT}-networks}.
\newblock In \emph{Proceedings of the 2019 Conference on Empirical Methods in Natural Language Processing and the 9th International Joint Conference on Natural Language Processing (EMNLP-IJCNLP)}, pages 3982--3992, Hong Kong, China. Association for Computational Linguistics.

\bibitem[{Song et~al.(2024)Song, Su, Shalyminov, Cai, and Mansour}]{song-etal-2024-finesure}
Hwanjun Song, Hang Su, Igor Shalyminov, Jason Cai, and Saab Mansour. 2024.
\newblock \href {https://doi.org/10.18653/v1/2024.acl-long.51} {{F}ine{S}ur{E}: Fine-grained summarization evaluation using {LLM}s}.
\newblock In \emph{Proceedings of the 62nd Annual Meeting of the Association for Computational Linguistics (Volume 1: Long Papers)}, pages 906--922, Bangkok, Thailand. Association for Computational Linguistics.

\bibitem[{Tang et~al.(2023)Tang, Wang, Goldsack, and Lin}]{tang-etal-2023-improving}
Chen Tang, Shun Wang, Tomas Goldsack, and Chenghua Lin. 2023.
\newblock \href {https://doi.org/10.18653/v1/2023.emnlp-main.40} {Improving biomedical abstractive summarisation with knowledge aggregation from citation papers}.
\newblock In \emph{Proceedings of the 2023 Conference on Empirical Methods in Natural Language Processing}, pages 606--618, Singapore. Association for Computational Linguistics.

\bibitem[{Team(2024)}]{qwen2.5}
Qwen Team. 2024.
\newblock \href {https://qwenlm.github.io/blog/qwen2.5/} {Qwen2.5: A party of foundation models}.

\bibitem[{Wu et~al.(2024)Wu, Huang, and Wei}]{wu2024mixture}
Xun Wu, Shaohan Huang, and Furu Wei. 2024.
\newblock \href {https://openreview.net/forum?id=uWvKBCYh4S} {Mixture of lo{RA} experts}.
\newblock In \emph{The Twelfth International Conference on Learning Representations}.

\bibitem[{Xu et~al.(2024)Xu, Lai, and Huang}]{xu2024meteora}
Jingwei Xu, Junyu Lai, and Yunpeng Huang. 2024.
\newblock Meteora: Multiple-tasks embedded lora for large language models.
\newblock \emph{arXiv preprint arXiv:2405.13053}.

\bibitem[{Xu et~al.(2016)Xu, Napoles, Pavlick, Chen, and {Callison-Burch}}]{xu-etal-2016-optimizing}
Wei Xu, Courtney Napoles, Ellie Pavlick, Quanze Chen, and Chris {Callison-Burch}. 2016.
\newblock \href {https://doi.org/10.1162/tacl_a_00107} {Optimizing statistical machine translation for text simplification}.
\newblock \emph{Transactions of the Association for Computational Linguistics}, 4:401--415.

\bibitem[{Yadav et~al.(2024)Yadav, Tam, Choshen, Raffel, and Bansal}]{yadav2024ties}
Prateek Yadav, Derek Tam, Leshem Choshen, Colin~A Raffel, and Mohit Bansal. 2024.
\newblock Ties-merging: Resolving interference when merging models.
\newblock \emph{Advances in Neural Information Processing Systems}, 36.

\bibitem[{Yu et~al.(2024)Yu, Yu, Yu, Huang, and Li}]{yu2024language}
Le~Yu, Bowen Yu, Haiyang Yu, Fei Huang, and Yongbin Li. 2024.
\newblock Language models are super mario: Absorbing abilities from homologous models as a free lunch.
\newblock In \emph{Forty-first International Conference on Machine Learning}.

\bibitem[{Zhang et~al.(2022)Zhang, Roller, Goyal, Artetxe, Chen, Chen, Dewan, Diab, Li, Lin et~al.}]{zhang2022opt}
Susan Zhang, Stephen Roller, Naman Goyal, Mikel Artetxe, Moya Chen, Shuohui Chen, Christopher Dewan, Mona Diab, Xian Li, Xi~Victoria Lin, et~al. 2022.
\newblock Opt: Open pre-trained transformer language models.
\newblock \emph{arXiv preprint arXiv:2205.01068}.

\bibitem[{Zhang* et~al.(2020)Zhang*, Kishore*, Wu*, Weinberger, and Artzi}]{bert-score}
Tianyi Zhang*, Varsha Kishore*, Felix Wu*, Kilian~Q. Weinberger, and Yoav Artzi. 2020.
\newblock {{BERTScore}}: {{Evaluating}} text generation with {{BERT}}.
\newblock In \emph{International Conference on Learning Representations}.

\bibitem[{Zhang and Li(2024)}]{zhang2024dlp}
Yuxuan Zhang and Ruizhe Li. 2024.
\newblock Dlp-lora: Efficient task-specific lora fusion with a dynamic, lightweight plugin for large language models.
\newblock \emph{arXiv preprint arXiv:2410.01497}.

\bibitem[{Zhao et~al.(2024{\natexlab{a}})Zhao, Zeng, Shi, and Zhou}]{zhao2024mosld}
Lulu Zhao, Weihao Zeng, Xiaofeng Shi, and Hua Zhou. 2024{\natexlab{a}}.
\newblock Mosld: An extremely parameter-efficient mixture-of-shared loras for multi-task learning.
\newblock \emph{arXiv preprint arXiv:2412.08946}.

\bibitem[{Zhao et~al.(2024{\natexlab{b}})Zhao, Gan, Wang, Zhou, Yang, Kuang, and Wu}]{zhao-etal-2024-loraretriever}
Ziyu Zhao, Leilei Gan, Guoyin Wang, Wangchunshu Zhou, Hongxia Yang, Kun Kuang, and Fei Wu. 2024{\natexlab{b}}.
\newblock \href {https://doi.org/10.18653/v1/2024.findings-acl.263} {{L}ora{R}etriever: Input-aware {L}o{RA} retrieval and composition for mixed tasks in the wild}.
\newblock In \emph{Findings of the Association for Computational Linguistics: ACL 2024}, pages 4447--4462, Bangkok, Thailand. Association for Computational Linguistics.

\bibitem[{Zhao et~al.(2024{\natexlab{c}})Zhao, Shen, Zhu, Li, Su, Wang, Kuang, and Wu}]{zhao2024merging}
Ziyu Zhao, Tao Shen, Didi Zhu, Zexi Li, Jing Su, Xuwu Wang, Kun Kuang, and Fei Wu. 2024{\natexlab{c}}.
\newblock Merging loras like playing lego: Pushing the modularity of lora to extremes through rank-wise clustering.
\newblock \emph{arXiv preprint arXiv:2409.16167}.

\bibitem[{Zheng et~al.(2024)Zheng, Zhang, Zhang, Ye, Luo, Feng, and Ma}]{zheng2024llamafactory}
Yaowei Zheng, Richong Zhang, Junhao Zhang, Yanhan Ye, Zheyan Luo, Zhangchi Feng, and Yongqiang Ma. 2024.
\newblock \href {http://arxiv.org/abs/2403.13372} {Llamafactory: Unified efficient fine-tuning of 100+ language models}.
\newblock In \emph{Proceedings of the 62nd Annual Meeting of the Association for Computational Linguistics (Volume 3: System Demonstrations)}, Bangkok, Thailand. Association for Computational Linguistics.

\bibitem[{Zhou and Bhat(2021)}]{zhou2021paraphrase}
Jianing Zhou and Suma Bhat. 2021.
\newblock Paraphrase generation: A survey of the state of the art.
\newblock In \emph{Proceedings of the 2021 conference on empirical methods in natural language processing}, pages 5075--5086.

\end{thebibliography}

\appendix

\section{More Related Works}
\begin{table*}[h]
\small
\centering
\caption{An example of the differences among text paraphrasing, text simplification, lay summarization and lay paraphrasing. The example input text is from VTechAGP dataset and results are generated by ChatGPT.}
\label{tab:data_example}
\begin{tabular}{|p{15cm}|}
\hline
\textbf{Input Text} \\
\hline
Thin filament pyrometry is a proven technique used to measure flame temperature by capturing the spectral radiance produced by the immersion of silicon carbide filaments in a hot gas environment. In this study a commercially available CMOS color camera was used, and the spectral response of each color channel was integrated with respect to the assumed graybody radiation spectrum to form a look up table between color ratio and temperature. Interpolated filament temperatures are then corrected for radiation losses via an energy balance to determine the flame temperature. Verification of the technique was performed on the Holthuis and Associates Flat Flame Burner, formerly known as the Mckenna Burner, and the results are directly compared to literature values measured on a similar burner. The results are also supported by radiation corrected measurements taken using a type B thermocouple on the same burner setup. An error propagation analysis was performed to determine which factors contribute the most to the final measurement uncertainty and confidence intervals are calculated for the results. Uncertainty values for a single point measurement were determined to be between ±15 and ±50 K depending on the color ratio and the total uncertainty associated with day-to-day changes in the measurement setup was found to be ±55 K.   \\
\hline
\textbf{Text Paraphrasing} \\
\hline
Thin filament pyrometry is an established method for measuring flame temperatures by analyzing the spectral radiance emitted when silicon carbide filaments are immersed in a hot gas. In this study, a commercially available CMOS color camera was utilized, and the spectral response of each color channel was integrated with the assumed graybody radiation spectrum to create a lookup table correlating color ratio to temperature. The filament temperatures obtained through interpolation were adjusted for radiation losses using an energy balance to determine the flame temperature. The method was validated using the Holthuis and Associates Flat Flame Burner (previously known as the Mckenna Burner), and the results were directly compared with literature values from similar burners. Additionally, radiation-corrected measurements from a type B thermocouple on the same setup supported the findings. An error propagation analysis identified the key contributors to measurement uncertainty, and confidence intervals were calculated. The uncertainty for a single measurement ranged from ±15 to ±50 K, depending on the color ratio, while day-to-day variations in the setup introduced an uncertainty of ±55 K.\\
\hline
\textbf{Text Simplification} \\
\hline
Thin filament pyrometry is a reliable method for measuring flame temperature by analyzing the light emitted when silicon carbide filaments are placed in hot gas. In this study, a CMOS color camera was used, and a table was created to link color ratios to temperature based on the assumed radiation spectrum. Filament temperatures were adjusted for radiation losses to calculate the flame temperature. The method was tested on a Flat Flame Burner, also known as the Mckenna Burner, and the results matched previously published data. Additional tests using a type B thermocouple confirmed the findings. An analysis of errors showed which factors affected measurement accuracy the most, with uncertainties ranging from ±15 to ±50 K for single measurements and ±55 K for variations in the setup over time.
\\
\hline
\textbf{Lay Summarization} \\
\hline
Thin filament pyrometry is a technique used to measure how hot a flame is by analyzing the light given off by a small filament placed in the flame. In this study, researchers used a camera to measure light colors and linked these to temperatures using a pre-made chart. They adjusted the results to account for energy lost as heat and tested the method on a standard burner. The findings matched results from other studies and were confirmed with another temperature-measuring tool. The researchers also identified factors that affect accuracy and found that temperature measurements could vary by about 15 to 55 K due to setup changes.
\\
\hline
\textbf{Lay Paraphrasing} \\
\hline
Thin filament pyrometry is a method used to measure flame temperatures by analyzing the light emitted from a small silicon carbide filament placed in hot gas. In this study, researchers used a regular CMOS color camera to capture light colors and created a chart to link color ratios to temperatures based on expected radiation patterns. They adjusted the temperature measurements to account for heat loss and tested their method on a standard burner setup, known as the Flat Flame Burner or Mckenna Burner. The results matched previous studies and were confirmed with another tool called a type B thermocouple. The researchers also examined what factors caused measurement errors and calculated that single temperature readings could vary by 15 to 50 K, while changes in the setup added an uncertainty of about 55 K.
\\
\hline
\end{tabular}

\end{table*}

Text paraphrasing aims at rewriting text by different words or sentence structures while keeping the original meaning~\cite{zhou2021paraphrase}.
Text simplification modifies syntax and lexicon to improve the understandability of language for young readers~\cite{al2021automated}.
Lay summarization involves generating a simplified summary of a technical or specialist text that
is suitable for a non-expert audience~\cite{goldsack-etal-2024-overview, giannouris-etal-2024-plain}.
Lay paraphrasing focuses on rewriting the text written from the technical experts to the general audience without specialized domain knowledge~\cite{cheng-etal-2025-vtechagp, guo2024retrieval}. 
Table~\ref{tab:data_example} shows examples of differences for the above four tasks.

\section{Dataset Analysis}
\label{sec:data_appendix}
In our experiments, we evaluate our model over five widely used public datasets: PLOS~\cite{goldsack-etal-2022-making}, eLife~\cite{goldsack-etal-2022-making}, CELLS~\cite{guo2024retrieval}, SciTechNews~\cite{cardenas-etal-2023-dont}, and VTechAGP~\cite{cheng-etal-2025-vtechagp}. We followed the original training, validation and testing splitting for PLOS, eLife, CELLS and VTechAGP. For SciTechNew, because the training set does not have abstract pairs while the validation and testing sets have the scientific and non-technical paraphraph pairs, we randomly resplit the dataset from original validation and testing sets into training, validation and testing sets by 0.8:0.1:0.1.
In the following, we describe each dataset in detail.
\begin{itemize}
    \item PLOS~\cite{goldsack-etal-2022-making} consists of abstracts from biomedical articles paired with non-technical lay summariesin science and medicine domain sourced from the peer-reviewed journals of The Public Library of Science publisher.
    \item eLife~\cite{goldsack-etal-2022-making} dataset contains scientific abstracts paired with non-technical lay summaries in the field of biomedical and life sciences derived from an open-access peer-reviewed journal.
    \item CELLS~\cite{guo2024retrieval} is the paragraph-paired dataset of scientific abstracts and expert-authored plain language summaries in the biomedicine field derived from biomedical journals for the lay language generation task. 
    \item SciTechNews~\cite{cardenas-etal-2023-dont} is a text-to-text science journalism dataset consisting of scientific papers paired with their corresponding press release snippet mined from ACM TechNews about scientific achievements and technology.
    \item VTechAGP~\cite{cheng-etal-2025-vtechagp} is an academic-to-general-audience text paraphrase dataset derived from electronic theses and dissertations across eight different colleges sourced from Virginia Tech Graduate School and Digital Libraries.
\end{itemize}

Following~\cite{cheng-etal-2025-vtechagp}, we report some basic dataset statistics in Table~\ref{tab:dataset}, including the number of documents for each public dataset, the average number of sentences for each document, and the average sentence length for the source and target texts, respectively. We also report Flesch-Kincaid Grade Level (FKGL) and Dale-Chall Readability Score (DCRS) follows~\cite{goldsack-etal-2022-making}.
Both FKGL and DCRS estimate the U.S. grade level required to comprehend a given text. FKGL calculates this based on the total count of sentences, words, and syllables in the text. In contrast, DCRS evaluates readability by considering the average sentence length and the number of familiar words, referencing a lookup table of the 3,000 most commonly used English words.

\section{Ablation Study}~\label{sec:distance}
\begin{table*}[]
\small
\centering
\tabcolsep=0.13cm
\caption{Full ablation study over Sci-LoRA components across 12 domains in all five datasets. (Part I)}
\label{tab:ablation_all}
\begin{tabular}{lcccccccccc}
\hline
            & s-BLEU         & d-BLEU         & BERTScore      & BLONDE         & ROUGE1         & ROUGE2         & METEOR         & COMET          & SARI           & FRES           \\ \hline
            & \multicolumn{10}{c}{CELLS}                                                                                                                                              \\ \hline
Pre-trained & 2.34           & 5.41           & 81.80          & 16.83          & 37.84          & 9.67           & 32.79          & \textbf{81.55} & 40.32          & 34.66          \\
Multi-LoRAs & 3.11           & 9.51           & 82.35          & \textbf{19.09} & 40.55          & 11.95          & 29.84          & 78.77          & 40.43          & 43.53          \\
Single LoRA & 3.07           & 9.26           & 82.36          & 18.80          & 40.55          & 11.92          & 30.40          & 78.95          & 40.38          & 43.43          \\
AWG\textsubscript{Random}           & 2.48           & 6.83           & 81.33          & 14.37          & 36.85          & 9.92           & 27.78          & 77.82          & 40.16          & 43.63          \\
AWG\textsubscript{K-Means}           & 2.41           & 6.58           & 81.30          & 15.26          & 36.74          & 9.79           & 27.99          & 78.11          & 40.24          & 43.73          \\
AWG\textsubscript{Contrastive}           & 2.56           & 7.09           & 81.49          & 16.42          & 37.38          & 10.08          & 27.98          & 77.92          & 40.33          & 43.53          \\
w/o Fusion  & 2.09           & 6.24           & 80.66          & 15.71          & 37.54          & 9.88           & 27.23          & 77.59          & 40.86          & 44.14          \\
Sci-LoRA    & \textbf{3.99}  & \textbf{11.15} & \textbf{83.00} & 18.12          & \textbf{43.10} & \textbf{14.02} & \textbf{32.29} & 79.10          & \textbf{41.15} & \textbf{44.36} \\ \hline
            & \multicolumn{10}{c}{PLOS}                                                                                                                                               \\ \hline
Pre-trained & 2.44           & 6.18           & 82.44          & 17.80          & 40.65          & 10.67          & 33.12          & \textbf{81.68} & 40.03          & 34.36          \\
Multi-LoRAs & 3.09           & 10.15          & 82.68          & 19.58          & 42.69          & 13.25          & 31.78          & 79.88          & 40.24          & 43.43          \\
Single LoRA & 3.23           & 10.18          & 82.70          & \textbf{21.46} & 42.75          & 13.25          & 31.88          & 79.90          & \textbf{40.08} & 43.43          \\
AWG\textsubscript{Random}           & 2.23           & 6.24           & 81.13          & 14.63          & 37.37          & 10.00          & 28.88          & 78.75          & 39.46          & 45.25          \\
AWG\textsubscript{K-Means}           & 2.35           & 6.38           & 81.20          & 13.90          & 37.45          & 10.13          & 29.23          & 78.95          & 39.62          & \textbf{45.46} \\
AWG\textsubscript{Contrastive}           & 2.39           & 6.57           & 81.41          & 16.64          & 37.97          & 10.22          & 28.54          & 78.66          & 39.62          & 44.95          \\
w/o Fusion  & 2.52           & 6.25           & 80.85          & 14.56          & 42.32          & 10.97          & 33.58          & 79.53          & 40.03          & 44.24          \\
Sci-LoRA    & \textbf{4.06}  & \textbf{12.43} & \textbf{83.35} & 16.81          & \textbf{45.59} & \textbf{15.89} & \textbf{35.20} & 80.29          & 40.07          & 44.36          \\ \hline
            & \multicolumn{10}{c}{eLife}                                                                                                                                              \\ \hline
Pre-trained & 0.85           & 3.17           & 80.52          & \textbf{5.00}  & 36.19          & 8.27           & 20.36          & 79.44          & 42.67          & 42.51          \\
Multi-LoRAs & 1.19           & 3.99           & 80.48          & 4.04           & 37.37          & 9.05           & 21.31          & 78.84          & 43.84          & 53.21          \\
Single LoRA & \textbf{1.22}  & 3.98           & 80.72          & 4.97           & 37.57          & 9.23           & 21.21          & 79.09          & 44.03          & 53.41          \\
AWG\textsubscript{Random}           & 1.12           & 6.25           & 81.24          & 4.36           & 41.95          & 10.10          & 27.66          & 81.82          & 47.39          & \textbf{55.13} \\
AWG\textsubscript{K-Means}          & 1.17           & 6.52           & 81.39          & 4.39           & 42.38          & 10.31          & 27.92          & 82.06          & 47.63          & 54.93          \\
AWG\textsubscript{Contrastive}           & 1.19           & \textbf{6.58}  & \textbf{81.62} & 4.89           & 42.22          & 10.41          & 28.40          & 82.54          & \textbf{47.80} & 54.83          \\
w/o Fusion  & 0.93           & 3.11           & 81.13          & 4.06           & 38.50          & 9.28           & 23.07          & 81.76          & 44.11          & 54.35          \\
Sci-LoRA    & \textbf{1.22}  & 6.09           & 81.40          & 4.99           & \textbf{42.59} & \textbf{11.31} & \textbf{28.98} & \textbf{82.95} & 47.64          & 54.93          \\ \hline
            & \multicolumn{10}{c}{SciTechNews}                                                                                                                                        \\ \hline
Pre-trained & 0.95           & 2.82           & 77.37          & 9.07           & 30.30          & 5.10           & \textbf{24.43} & \textbf{76.29} & 39.15          & 34.76          \\
Multi-LoRAs & 1.84           & 4.09           & 78.00          & 8.74           & 31.93          & 5.92           & 22.89          & 74.35          & 43.35          & 41.29          \\
Single LoRA & 1.58           & 4.26           & 77.90          & 9.10           & 31.85          & 6.23           & 22.90          & 74.32          & 43.02          & \textbf{42.11} \\
AWG\textsubscript{Random}           & 2.03           & 4.09           & 78.01          & 9.04           & 31.60          & 6.11           & 21.79          & 73.64          & 43.66          & 41.09          \\
AWG\textsubscript{K-Means}          & 2.07           & 4.11           & 78.83          & 9.60           & 31.73          & 6.79           & 22.99          & 73.81          & 43.52          & 41.19          \\
AWG\textsubscript{Contrastive}           & 2.19           & 4.17           & \textbf{78.88} & 9.89           & 32.29          & 6.88           & 22.21          & 75.15          & 43.42          & 41.29          \\
w/o Fusion  & 1.18           & 4.03           & 78.30          & 8.91           & 30.39          & 6.22           & 23.09          & 74.36          & 42.68          & 38.26          \\
Sci-LoRA    & \textbf{2.20}  & \textbf{4.61}  & 78.82          & \textbf{10.13} & \textbf{32.68} & \textbf{6.90}  & 23.87          & 76.00          & \textbf{43.76} & 41.38          \\ \hline
            & \multicolumn{10}{c}{College of Agriculture and Life Sciences (VTechAGP)}                                                                                                \\ \hline
Pre-trained & 3.92           & 13.04          & 84.15          & 10.71          & 48.78          & 16.96          & 35.58          & \textbf{85.46} & 36.91          & 34.97          \\
Multi-LoRAs & 8.73           & 27.00          & 84.97          & 16.05          & 52.49          & 27.97          & 41.29          & 82.88          & 40.12          & 34.76          \\
Single LoRA & 8.95           & 25.55          & 84.98          & 36.02          & 52.90          & 27.50          & 40.88          & 83.42          & 40.14          & 35.07          \\
AWG\textsubscript{Random}           & 9.01           & 26.61          & 84.77          & 36.92          & 52.52          & 26.97          & 40.51          & 82.65          & 39.72          & 34.36          \\
AWG\textsubscript{K-Means}           & 9.26           & 26.67          & 84.92          & 35.90          & 53.10          & 27.30          & 40.96          & 82.84          & 40.45          & 42.82          \\
AWG\textsubscript{Contrastive}           & 9.91           & 26.83          & 84.83          & 36.41          & 53.20          & 27.93          & 40.91          & 82.97          & 40.72          & 42.56          \\
w/o Fusion  & 6.18           & 24.67          & 84.47          & 22.22          & 52.74          & 27.39          & 40.56          & 82.66          & 40.38          & 40.86          \\
Sci-LoRA    & \textbf{10.21} & \textbf{31.03} & \textbf{86.01} & \textbf{36.99} & \textbf{56.90} & \textbf{32.16} & \textbf{45.51} & 83.79          & \textbf{41.32} & \textbf{44.86} \\ \hline
            & \multicolumn{10}{c}{College of Architecture, Arts, and Design (VTechAGP)}                                                                                               \\ \hline
Pre-trained & 5.63           & 12.69          & 81.97          & 27.56          & 42.46          & 13.87          & 33.98          & \textbf{80.05} & 34.64          & 35.17          \\
Multi-LoRAs & 19.78          & 34.39          & 83.30          & 46.43          & 47.83          & 26.51          & 41.77          & 77.20          & 40.76          & 42.41          \\
Single LoRA & 13.43          & 28.71          & 82.55          & 40.64          & 46.51          & 23.37          & 39.85          & 76.06          & 39.39          & \textbf{51.89} \\
AWG\textsubscript{Random}           & 23.08          & 34.74          & 83.46          & 48.14          & 48.81          & 26.16          & 40.48          & 77.30          & 41.58          & 42.41          \\
AWG\textsubscript{K-Means}          & \textbf{23.17} & 37.08          & 83.72          & 50.02          & 51.13          & 28.28          & 43.91          & 77.83          & 43.56          & 42.31          \\
AWG\textsubscript{Contrastive}           & 18.06          & 38.62          & 84.03          & 50.00          & 52.07          & 30.86          & 44.89          & 77.97          & 44.00          & 41.29          \\
w/o Fusion  & 14.86          & 31.22          & 81.59          & 44.02          & 50.82          & 22.62          & 41.42          & 78.59          & 43.63          & 44.03          \\
Sci-LoRA    & 18.38          & \textbf{38.97} & \textbf{84.37} & \textbf{50.99} & \textbf{52.69} & \textbf{30.98} & \textbf{46.71} & 78.02          & \textbf{44.26} & 41.80          \\ \hline
\end{tabular}
\end{table*}

\begin{table*}[]
\small
\centering
\tabcolsep=0.13cm
\caption{Full ablation study over Sci-LoRA components across 12 domains in all five datasets. (Part II)}
\label{tab:ablation_all_2}
\begin{tabular}{lcccccccccc}
\hline
            & s-BLEU         & d-BLEU         & BERTScore      & BLONDE         & ROUGE1         & ROUGE2         & METEOR         & COMET          & SARI           & FRES           \\ \hline
            & \multicolumn{10}{c}{College of Engineering (VTechAGP)}                                                                                                                  \\ \hline
Pre-trained & 3.57           & 10.74          & 82.88          & 6.39           & 44.78          & 14.45          & 34.85          & \textbf{83.84} & 35.85          & 26.20          \\
Multi-LoRAs & 9.72           & 24.45          & 83.57          & 9.52           & 48.39          & 23.38          & 39.08          & 82.02          & 38.73          & 32.63          \\
Single LoRA & 9.72           & 24.40          & 83.16          & 9.27           & 47.54          & 23.20          & 38.47          & 81.22          & 38.74          & 32.43          \\
AWG\textsubscript{Random}           & 11.13          & 25.18          & 83.72          & 9.44           & 48.74          & 24.60          & 39.79          & 81.69          & 40.55          & 32.43          \\
AWG\textsubscript{K-Means}           & 11.60          & 25.77          & 83.59          & 9.74           & 48.74          & 24.87          & 39.99          & 81.56          & 41.32          & 32.33          \\
AWG\textsubscript{Contrastive}           & \textbf{11.69} & 25.95          & 83.92          & 9.94           & 49.54          & 24.94          & 40.59          & 82.25          & \textbf{41.64} & 33.33          \\
w/o Fusion  & 9.33           & 23.74          & 83.86          & 9.69           & 46.52          & 22.16          & 38.78          & 82.81          & 38.48          & \textbf{33.85} \\
Sci-LoRA    & \textbf{11.69} & \textbf{28.31} & \textbf{84.30} & \textbf{10.28} & \textbf{51.45} & \textbf{27.62} & \textbf{42.73} & 82.90          & \textbf{41.64} & 33.82          \\ \hline
            & \multicolumn{10}{c}{College of Liberal Arts and Human Sciences (VTechAGP)}                                                                                              \\ \hline
Pre-trained & 6.95           & 14.62          & 84.35          & 30.19          & 48.16          & 19.00          & 39.44          & \textbf{83.42} & 33.34          & \textbf{33.34} \\
Multi-LoRAs & 20.38          & 37.63          & 86.30          & 45.29          & 56.44          & 36.19          & 49.57          & 81.86          & 40.75          & 30.80          \\
Single LoRA & 19.82          & 36.59          & 86.24          & 46.14          & 56.03          & 35.59          & 49.26          & 81.86          & 38.98          & 30.50          \\
AWG\textsubscript{Random}           & 23.26          & 35.87          & 86.35          & 45.44          & 56.20          & 36.04          & 49.16          & 81.89          & 40.20          & 30.20          \\
AWG\textsubscript{K-Means}           & 23.44          & 38.47          & 86.66          & \textbf{48.92} & 57.45          & 37.46          & 51.10          & 82.20          & 42.26          & 30.09          \\
AWG\textsubscript{Contrastive}           & 23.92          & 38.71          & 86.62          & 48.83          & 57.93          & 37.95          & 51.41          & 82.56          & 42.56          & 30.90          \\
w/o Fusion  & 18.07          & 37.69          & 86.31          & 42.23          & 56.96          & 35.41          & 49.37          & 82.44          & 39.72          & 32.73          \\
Sci-LoRA    & \textbf{23.99} & \textbf{40.33} & \textbf{87.25} & 48.32          & \textbf{59.59} & \textbf{40.92} & \textbf{53.37} & 82.62          & \textbf{42.88} & 32.48          \\ \hline
            & \multicolumn{10}{c}{College of Natural Resources and Environment (VTechAGP)}                                                                                            \\ \hline
Pre-trained & 3.96           & 12.24          & 84.34          & 23.64          & 49.13          & 16.96          & 37.53          & \textbf{85.81} & 35.85          & 35.27          \\
Multi-LoRAs & 11.12          & 26.78          & 85.11          & 35.63          & 53.79          & 29.24          & 43.89          & 83.25          & 40.91          & 42.82          \\
Single LoRA & 7.55           & 24.42          & 84.75          & 34.18          & 51.67          & 26.53          & 41.31          & 83.23          & 38.49          & 42.82          \\
AWG\textsubscript{Random}           & 11.66          & 27.60          & 85.48          & 38.64          & 54.67          & 29.93          & 45.09          & 84.17          & 40.82          & 41.50          \\
AWG\textsubscript{K-Means}           & 11.76          & 28.07          & 85.51          & 38.43          & 54.96          & 30.21          & 45.14          & 84.08          & 40.72          & 41.80          \\
AWG\textsubscript{Contrastive}           & 11.93          & 28.80          & 85.71          & 38.64          & 54.93          & 30.30          & 45.80          & 84.35          & 40.93          & 42.90          \\
w/o Fusion  & 11.44          & 27.04          & 85.01          & 35.10          & 52.36          & 29.95          & 42.81          & 83.05          & 39.87          & \textbf{43.22} \\
Sci-LoRA    & \textbf{13.84} & \textbf{29.61} & \textbf{86.18} & \textbf{40.41} & \textbf{57.30} & \textbf{33.19} & \textbf{47.59} & 84.66          & \textbf{41.77} & 42.80          \\ \hline
            & \multicolumn{10}{c}{College of Science (VTechAGP)}                                                                                                                      \\ \hline
Pre-trained & 3.50           & 10.30          & 82.33          & 6.20           & 43.32          & 13.45          & 33.89          & \textbf{82.70} & 37.26          & \textbf{34.36} \\
Multi-LoRAs & \textbf{11.07} & 20.96          & 82.91          & 8.57           & 46.05          & 21.45          & 37.15          & 79.90          & 38.61          & 33.65          \\
Single LoRA & 8.22           & 20.56          & 82.85          & 8.47           & 45.58          & 20.59          & 36.20          & 80.33          & 38.74          & 33.34          \\
AWG\textsubscript{Random}           & 8.57           & 21.19          & 82.78          & 8.49           & 44.86          & 20.48          & 35.43          & 80.18          & 38.27          & 33.14          \\
AWG\textsubscript{K-Means}           & 8.92           & 22.31          & 82.92          & 8.67           & 46.03          & 22.05          & 37.22          & 80.03          & 39.17          & 33.65          \\
AWG\textsubscript{Contrastive}           & 8.93           & 22.87          & 82.94          & 13.40          & 46.13          & 22.23          & 37.91          & 80.11          & 39.96          & 33.40          \\
w/o Fusion  & 8.15           & 20.56          & 82.64          & 8.46           & 45.19          & 20.81          & 36.73          & 80.85          & 38.04          & 33.95          \\
Sci-LoRA    & 9.86           & \textbf{23.31} & \textbf{83.51} & \textbf{14.75} & \textbf{48.62} & \textbf{24.26} & \textbf{40.18} & 80.87          & \textbf{40.60} & 33.44          \\ \hline
            & \multicolumn{10}{c}{Pamplin College of Business (VTechAGP)}                                                                                                             \\ \hline
Pre-trained & 4.13           & 12.78          & 84.27          & 28.00          & 47.04          & 17.90          & 36.78          & 85.38          & 38.38          & \textbf{23.97} \\
Multi-LoRAs & 16.03          & 29.67          & 85.22          & 43.33          & 52.17          & 29.32          & 43.09          & 83.56          & 42.96          & 21.43          \\
Single LoRA & 15.49          & 23.55          & 84.11          & 34.22          & 47.53          & 23.35          & 37.67          & 82.60          & 39.15          & 23.04          \\
AWG\textsubscript{Random}           & 15.29          & 26.85          & 85.41          & 41.07          & 53.11          & 27.92          & 43.90          & 83.48          & 41.07          & 21.62          \\
AWG\textsubscript{K-Means}           & 15.82          & 28.69          & 85.71          & 42.39          & 54.02          & 28.85          & 44.81          & 84.14          & 42.07          & 21.33          \\
AWG\textsubscript{Contrastive}           & \textbf{16.09} & 29.38          & 85.90          & 43.39          & 54.78          & 28.86          & 44.58          & 84.08          & 42.92          & 21.84          \\
w/o Fusion  & 13.10          & 22.31          & 84.57          & 35.42          & 53.30          & 23.70          & 42.51          & 84.43          & 40.31          & 23.62          \\
Sci-LoRA    & 15.76          & \textbf{32.86} & \textbf{86.51} & \textbf{46.58} & \textbf{56.90} & \textbf{34.50} & \textbf{48.36} & \textbf{85.40} & \textbf{45.08} & 22.53          \\ \hline
            & \multicolumn{10}{c}{Virginia Maryland College of Veterinary Medicine (VTechAGP)}                                                                                        \\ \hline
Pre-trained & 3.57           & 11.12          & 83.49          & 22.99          & 46.25          & 15.08          & 34.88          & \textbf{84.83} & 36.98          & \textbf{43.12} \\
Multi-LoRAs & 9.80           & 24.33          & 83.99          & 31.49          & 49.37          & 22.87          & 37.83          & 82.68          & 39.81          & 33.14          \\
Single LoRA & 9.32           & 21.78          & 84.37          & 31.32          & 48.60          & 22.02          & 37.56          & 83.58          & 41.13          & 42.31          \\
AWG\textsubscript{Random}           & 10.92          & 22.38          & 84.20          & 33.06          & 47.25          & 22.47          & 35.17          & 81.91          & 36.61          & 32.53          \\
AWG\textsubscript{K-Means}          & 11.01          & 29.11          & 85.11          & \textbf{39.98} & 52.65          & 27.69          & 42.22          & 84.35          & 42.17          & 33.24          \\
AWG\textsubscript{Contrastive}           & \textbf{11.81} & 29.26          & 85.48          & 39.32          & 52.95          & 27.97          & 42.78          & 84.58          & 42.20          & 33.94          \\
w/o Fusion  & 9.80           & 22.80          & 84.00          & 28.83          & 48.16          & 22.76          & 39.58          & 81.48          & 40.26          & 33.95          \\
Sci-LoRA    & 11.32          & \textbf{29.55} & \textbf{85.90} & 39.09          & \textbf{53.80} & \textbf{28.24} & \textbf{44.00} & 84.67          & \textbf{42.45} & 34.12          \\ \hline
\end{tabular}
\end{table*}

Table~\ref{tab:ablation_all} and Table~\ref{tab:ablation_all_2} present the complete results of the ablation study on Sci-LoRA's components across 12 scientific domains over five distinct datasets. From both tables, we observe that fine-tuning pre-trained models with LoRA (both Multi-LoRAs and Single LoRA) significantly enhances performance. In most cases, fine-tuning a dedicated LoRA for a specific domain yields better results than using a single LoRA across all domains. However, there are notable exceptions. For instance, on the eLife and SciTechNews datasets, a single LoRA outperforms multiple domain-specific LoRAs. This is likely due to the relatively small training dataset sizes for these domains, making it challenging to learn robust representations with limited data. Additionally, some topics in the eLife dataset are closely related to the biomedical domain, which aligns with CELLS—the largest training dataset. As a result, fine-tuning a single LoRA across all datasets allows it to leverage knowledge from CELLS, leading to improved performance.

All components of Sci-LoRA—the k-means-based adapter weight generator, the contrastive-learning-trained text encoder, and the dynamic fusion module—play crucial roles in its effectiveness. When integrated, these components enable Sci-LoRA to consistently outperform other methods across multiple key metrics. Notably, it achieves the highest scores in sentence BLEU, document BLEU, BERTScore, ROUGE-1, ROUGE-2, METEOR, and SARI across most domains, demonstrating that domain-specific fine-tuning significantly enhances text generation quality.
One notable observation is that for COMET, an end-to-end evaluation metric, pre-trained models consistently yield the best performance. Further exploration into improving Sci-LoRA's end-to-end performance remains an avenue for future research.

To analyze the impact of the fine-tuned text encoder using contrastive learning as discussed in Sec.~\ref{sec:AWG}, we present t-SNE visualizations of the text embeddings generated by the encoder both before and after contrastive learning.
The embeddings are from five datasets: CELLS, eLife, SciTechNews, PLOS, and ENG in VTechAGP. For better demonstration, we select the largest domain ENG from VTechAGP for visualization.
As shown in Figure~\ref{fig:visualization}, the data points corresponding to different domains (CELLs, eLife, SciTechNews, PLOS, ENG) are intermingled. 
There is no clear separation between the domains, indicating that the embeddings produced by the pre-trained text encoder (without contrastive learning) do not capture domain-specific features effectively. 
The central point (average embedding) for each domain is shown as a star in Figure~\ref{fig:visualization}. 
The closeness of points from different domains indicates that embeddings for different domains are overly similar, making it difficult to differentiate between them.

After using contrastive learning for fine-tuning the text encoder as discussed in Sec.~\ref{sec:AWG}, the embeddings exhibit much clearer separations between domains. 
Each domain forms a distinct cluster, demonstrating that the contrastive learning process successfully pushed embeddings from different domains apart while pulling embeddings within the same domain closer.
Note that embeddings of PLOS dataset (red points) and embeddings of CELLS dataset (blue points) are still intermingled after contrastive learning. 
This is because as discussed in CELLS~\cite{guo2024retrieval}, CELLS dataset has an overlap with PLOS dataset~\cite{goldsack-etal-2022-making}, and text in both datasets are mainly from biomedical domain. 
In addition, there still exist text embeddings from different domains that are intermingled (overlapped data points with different colors).
This is because the data may inherently have overlapping features across domains. 
For example, some text samples may contain content relevant to both the green cluster (eLife domain) and the orange cluster (SciTechNews domain), causing their embeddings to appear closer in the high-dimensional space.
This also validates that there may be semantic similarities between certain domains. 
For instance, both SciTechNews and eLife could have overlapping topics such as scientific advancements, which makes their embeddings similar and distance closer.

\begin{figure*}[htp] 
 \center{\includegraphics[height=18cm,width=\textwidth]{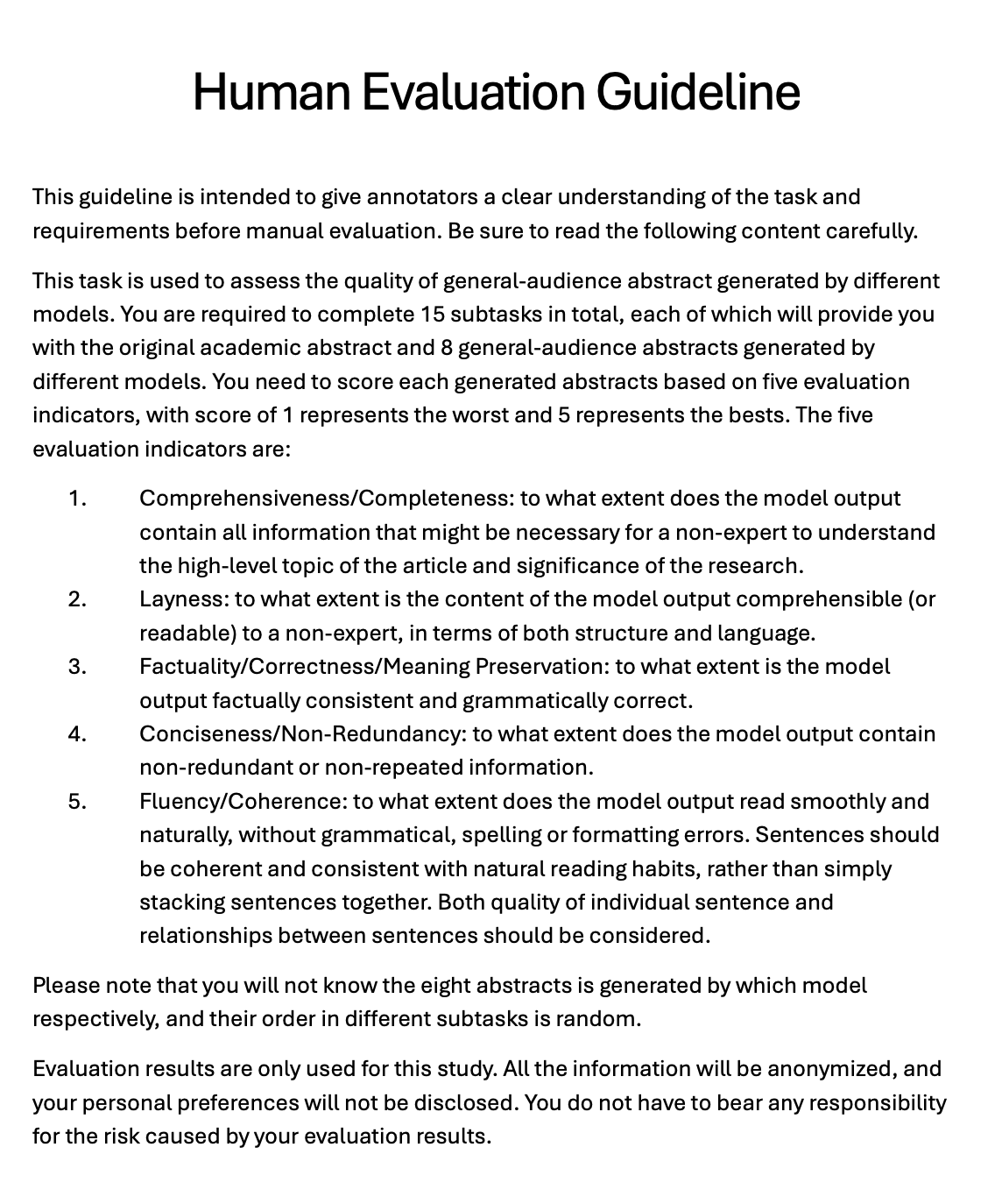}}
 \caption{\label{fig:human} Human evaluation guideline.}
 \end{figure*}

\end{document}